\documentclass[12pt]{iopart}
\usepackage{graphicx}
\usepackage{amssymb}
\usepackage{amsfonts}
\usepackage{bm}

\usepackage{booktabs} 

\usepackage{scalerel,stackengine} 
\stackMath
\newcommand\reallywidehat[1]{%
\savestack{\tmpbox}{\stretchto{%
  \scaleto{%
    \scalerel*[\widthof{\ensuremath{#1}}]{\kern-.6pt\bigwedge\kern-.6pt}%
    {\rule[-\textheight/2]{1ex}{\textheight}}
  }{\textheight}%
}{0.5ex}}%
\stackon[1pt]{#1}{\tmpbox}%
}

\newtheorem{remark}{Remark}

\begin{document}

\title[Hyper-Molecules]{Hyper-Molecules: on the Representation and Recovery of Dynamical Structures, with Application to Flexible Macro-Molecular Structures in Cryo-EM}
\author{Roy R. Lederman, $^1$\footnote{Part of the work was done while at the Program in Applied and Computational Mathematics, Princeton University, Princeton, NJ},  Joakim And\'{e}n $^2$ and  Amit Singer$^3$ }

\address{$^1$ The Department of Statistics and Data Science, Yale University, New Haven, CT} \ead{roy.lederman@yale.edu }

\address{$^2$ Center for Computational Mathematics, Flatiron Institute, New York, NY} \ead{janden@flatironinstitute.org}

\address{$^3$ Department of Mathematics and Program in Applied and Computational Mathematics, Princeton University, Princeton, NJ} \ead{amits@math.princeton.edu}

\begin{abstract}

    Cryo-electron microscopy (cryo-EM), the subject of the 2017 Nobel Prize in Chemistry, is a technology for determining the 3-D structure of macromolecules from many noisy 2-D projections of instances of these macromolecules, whose orientations and positions are unknown. 
    The molecular structures are not rigid objects, but flexible objects involved in dynamical processes. 
    The different conformations are exhibited by different instances of the macromolecule observed in a cryo-EM experiment, each of which is recorded as a particle image. The range of conformations and the conformation of each particle are not known a priori; one of the great promises of cryo-EM is to map this conformation space.
    Remarkable progress has been made in determining rigid structures from homogeneous samples of molecules in spite of the unknown orientation of each particle image 
    and significant progress has been made in recovering a few distinct states from mixtures of rather distinct conformations, 
    but more complex heterogeneous samples remain a major challenge. 
    
    We introduce the ``hyper-molecule'' framework for modeling structures across different states of heterogeneous molecules,
    including continuums of states.
    The key idea behind this framework is representing heterogeneous macromolecules as high-dimensional objects, with the additional dimensions 
    representing the conformation space. This idea is then refined to model properties such as localized heterogeneity. 
    In addition, we introduce an algorithmic framework for recovering such maps of heterogeneous objects from experimental data
    using a Bayesian formulation of the problem and Markov chain Monte Carlo (MCMC) algorithms to address the computational challenges in recovering these high dimensional hyper-molecules.
    We demonstrate these ideas in a prototype applied to synthetic data.

\end{abstract}


\maketitle

\section{Introduction}

Cryo-electron microscopy (cryo-EM) 
is joining X-ray crystallography and nuclear magnetic resonance (NMR) as a technology for recovering high-resolution structures of biological molecules \cite{kuhlbrandt,smith2014beyond,cheng-trpv1,amunts2014structure,bartesaghi20152}. A typical study produces hundreds of thousands of extremely noisy images of individual particles where the orientation of each individual particle is unknown, giving rise to a massive computational and statistical challenge. 
Current algorithms (e.g., \cite{scheres-relion,brubaker,eman2,van2006four,de2013xmipp,frealign}) have been  successful in recovering remarkably high-resolution structures of static macromolecules in homogeneous samples  with little variability, and have also been rather successful in recovering structures from heterogeneous samples consisting of a small number of distinct different structures (referred to as discrete heterogeneity). 
Even in homogeneous cases, there is ongoing work on improving  resolution, and there are several open questions about validating the results and estimating the uncertainty in the solutions.

Structural variations are intrinsic to the function of many macromolecules. Molecular motors, ion pumps, receptors, ion channels, polymerases, ribosomes, and spliceosomes are some of the molecular machines for which conformational fluctuations are essential to function. As just one example, the reaction cycle of the molecular motor kinesin is seen to involve a combination of discrete states (i.e., bound kinesin monomers in different stages of ATP hydrolysis) and also a continuous motion in which one monomer ``strides'' ahead while it is tethered by a linker to its microtubule-bound companion \cite{liu2017structural}. As another example, fluctuations in the conformation of ligand-binding domains drive the response of neuronal glutamate receptors \cite{dolino2016conformational}.
While technologies like X-ray crystallography and NMR measure ensembles of particles, cryo-EM produces images of individual particles, 
and one of the great promises of cryo-EM is that these noisy images, depicting individual particles at unknown states viewed from unknown directions, could potentially be compiled into maps of the dynamical processes in which these macromolecules participate \cite{nogales2016development,glaeser2016good}.
This, in turn, would help uncover the functionality of these molecular machines.

Due to the difficulties in the analysis of heterogeneous samples, researchers attempt to purify homogeneous samples; in doing so they lose information about other states/ conformations. 
Alternatively, they model the macromolecules observed in heterogeneous samples as a small number of distinct macromolecules (e.g., \cite{scheres}); this approach overlooks relationships between states (e.g., similarity between different conformations  of the molecule) and leads to an impractical number of distinct objects when the variability is complex or when there is a continuum of states rather than distinct independent states.
Currently, the analysis of heterogeneous macromolecules often misses states, achieves limited resolution, or yields remarkably high-resolution static structures, from which hang ``blurry'' heterogeneous pieces that cannot be accurately recovered. 
 The study of heterogeneity is considered an open problem without a well-established solution (see the recent survey \cite{sorzano2019survey}); existing approaches often rely on assumptions such as small modes of perturbation or piece-wise rigidity. In other cases, they require a reliable alignment of images before the heterogeneity can be addressed.

In some ways, the heterogeneity problem in cryo-EM is an extreme case of related problems that appear in the analysis of other systems that exhibit some intrinsic variability, 
such as the imaging of the body of a patient in computed tomography (CT) while the patient breathes \cite{low2003method} (in this case, the viewing directions are known, and there are some indications for the state in the breathing cycle).

We introduce a new mathematical framework with a Bayesian formulation for describing and mapping continuous heterogeneity in macromolecules
and an algorithmic approach  for computing these heterogeneous structures which addresses some of the computational and statistical challenges.
We present a preliminary implementation of these frameworks and experimental results.
Ultimately, the goal of this line of work is to produce scalable computational tools for mapping complex heterogeneity in macromolecules.
One of the goals in this design is to allow the use of a wide range of models and solvers that would enable the user to encode prior knowledge about the specific macromolecule being studied.
For the implementation of these ideas, we envision software for modeling of complex heterogeneous molecules in computer code (or simpler interfaces for common templates) as differentiable components, analogous to deep neural network models.
The prototype presented in this paper to demonstrate these ideas is more modest in its capabilities and scalability.

We start with the question: What does it mean to recover a heterogeneous macromolecule compared to a homogeneous/rigid macromolecule?
We propose that this boils down  to the question of representing a heterogeneous macromolecule in all its states; 
in other words, a ``solution'' would allow us to view the macromolecule at any state in a user interface that would provide us with ``knobs'' that we could turn to observe the molecule transition between states through a continuum of states. 
Often, it is useful to have statistics of how populated the states are, along with the map of states.
We recall the representation of molecules as 3-D functions using a linear combination of 3-D basis functions:
\begin{equation}\label{eq:objgeneral}
  \mathcal{V}({\bm r}) = \sum_{k} a_{k} \psi_{k}({\bm r}) , 
\end{equation}
with spatial coordinates ${\bm r}$.
We generalize this representation to describe a heterogeneous macromolecule in all its states. This generalization, which we refer to as a ``hyper-molecule,'' is described as follows. 
In Section \ref{sec:hyperobj}, we propose a generic generalization of (\ref{eq:objgeneral}). We represent hyper-molecules as linear combinations of higher-dimensional basis functions $\tilde{\psi}_{q}$:
\begin{equation}\label{eq:hyperobjgeneral}
  \mathcal{V}({\bm r},{\bm \tau}) = \sum_{q} a_{q} \tilde{\psi}_{q}({\bm r},{\bm {\tau}}) , 
\end{equation}
where the new dimensions capture heterogeneity, so that ${\bm \tau}$ identifies a conformation, or a location in the map of states,
and the macromolecule at state/conformation ${\bm \tau}$ is the 3-D density function obtained by fixing ${\bm \tau}$ in $\mathcal{V}(\cdot,{\bm \tau})$.
In other words, we generalize the classic problem of ``estimating a homogeneous macromolecule'' to the problem of ``estimating a heterogeneous hyper-molecule,'' 
a single high-dimensional object that encodes all the conformations of the macromolecule together. The (possibly high-dimensional) variable ${\bm \tau}$ represents the map of states, 
or the ``knobs'' which a user would turn in order to transition between states.
Furthermore, we argue that hyper-molecules are not merely a way to express the solution of some computation:
the representation through a finite set of basis functions serves as a regularizer in the computational problem, much like band-limit assumptions
in many inverse problems, including the homogeneous case of cryo-EM. 
In particular, the high-dimensional basis functions, each supported on multiple states, impose relations between states and define a continuum of states.
This property distinguishes between hyper-molecules and a small set of independent macromolecules.
This mathematical model of heterogeneous macromolecular structures is accompanied by a Bayesian formulation of recovering hyper-molecules from data,
which is a generalization of the Bayesian formulation of cryo-EM that allows a continuum of states and addresses the relationships between states. 

Increasingly complex heterogeneity is formulated using increasingly higher-dimensional hyper-molecules.
However, in Section \ref{sec:dim} we find that these hyper-molecules can be ``too generic'': the natural generalization of traditional algorithms to 
recover very high-dimensional hyper-molecules requires impractically large datasets and computational resources.
We address these problems in the remaining subsections of Section \ref{sec:analysis} and in Section \ref{sec:alg}.

First, in Section \ref{sec:compositehyperobj}, we introduce ``{\em composite hyper-molecules},'' a generalization of hyper-molecules that
 capture additional properties of macromolecules often known to scientists or readily identifiable.
Specifically, a macromolecule can often be modeled as a sum of $M$ rigid and heterogeneous components $\mathcal{V}^{m}$, each with its own state ${\bm \tau}^m$.
The state determines not only the shape of the component, but also its position with respect to the other components through a function denoted by $f^m$:
\begin{equation}\label{eq:compositehyperobj:traj:intro}
  \mathcal{V}({\bm r},{\bm \tau}^1, {\bm \tau}^2, ..., {\bm \tau}^M) =  \sum_{m=1}^M \mathcal{V}^{m} ( f^m ({\bm r},{\bm \tau}^m),{\bm \tau}^m).
\end{equation}
In this case, ``recovering the heterogeneous macromolecule'' means recovering the coefficients that describe each individual component $\mathcal{V}^{m}$ of $\mathcal{V}$
and recovering the coefficients that describe the trajectory $f^m$ of each component.

Next, in Section \ref{sec:blackboxhyperobj}, we note that the Bayesian formulation of hyper-molecule does not rely on a  specific representation of the hyper-molecule
and it interacts with the model of the hyper-molecule mainly through the comparison of particle images with the hyper-molecule at certain viewing directions and states, and through priors on the hyper-molecule structure. Therefore, we may replace our proposed hyper-molecules and composite hyper-molecules with other models, having coefficients ${\bm \theta}$.
We would then have an algorithm which accesses a black-box function $\mathcal{V}[{\bm \theta}]({\bm r},{\bm \tau})$ and a prior $P({\bm \theta})$,
and updates the coefficients ${\bm \theta}$, the viewing directions, state variables and so on without explicit knowledge of the detailed in the model of $\mathcal{V}$.
This formulation, which separates the model and prior of the hyper-molecule from the algorithm allows users to define more elaborate models as needed in their application.

The high-dimensional nature of hyper-molecules leads to a computational challenge. Specifically, the main computational challenge in current software packages, such as RELION \cite{scheres-relion,kimanius,zivanov2018new} and cryoSPARC \cite{brubaker}, is that each iteration of the algorithms involves a comparison of each particle image to the current estimate of the molecule as viewed from any possible direction (despite  modifications that reduce the number of comparisons required in practice significantly).
In hyper-molecules, we add the high-dimensional state variable $\bm{\tau}$, so that the natural generalization of current algorithms would require comparison of each particle image to each possible molecule (i.e., the hyper-molecule at any possible state) at each possible viewing direction, increasing the computational complexity exponentially with the increase in dimensionality. 
The variability in the nature of the heterogeneity and models makes it more challenging to develop generic solutions for reducing the number of comparisons.
In Section \ref{sec:alg} we propose a framework based on Markov chain Monte Carlo (MCMC) algorithms to address some of the computational complexity. This framework, allows complex, flexible, programmable black-box models and bypasses the need for exhaustive searches in each iteration.

In Section \ref{sec:res}, we present a prototype which implements hyper-molecules and MCMC, and present results from experiments with synthetic data.
This prototype demonstrates the applicability of hyper-molecules, composite hyper-molecules and MCMC to the mapping of continuous heterogeneity. 
We are currently developing the next version, which will be more scalable and allow more general models of hyper-molecules.

 Some of the preliminary work leading to this paper is available in an earlier technical report \cite{lederman2017continuously}.

\section{Preliminaries}

The purpose of this section is to briefly review some of the technical tools used in this paper.
In addition, we present the cryo-EM problem and related work on the problem, and we formulate the mathematical and statistical models which we will generalize in the remainder of the paper.

 \begin{table}[h!]\caption{Table of Notation}
          \begin{center}
            \begin{tabular}{r c p{10cm} }
              \toprule
              $\mathcal{V}$         & ~ & three- or higher-dimensional function \\
              $ \hat{\mathcal{V}} $ & ~ &  the Fourier transform of $\mathcal{V}$ in spatial coordinates \\
              $ R {\bm r} $  & ~ &   the vector $\bm{r}$ rotated by $R$ \\
              $ R \mathcal{V} $ & ~ &  the function $\mathcal{V}$ rotated by $R$, so that $\left( R \mathcal{V} \right)(\bm{x}) = \mathcal{V}(R^{-1}\bm{x})$  \\
              
              \bottomrule
              ${\bm r}$ & ~ & bold fonts are used to emphasize that a certain variable may be a vector, not just a scalar, when this is not obvious from the context.  \\
              
              \bottomrule
            \end{tabular}
          \end{center}
          \label{tb:notation}
        \end{table}

\subsection{Representation of Functions}\label{sec:pre:representations}

        A function such as $f: \mathcal{X} \rightarrow \mathbb{R} $ can be represented in many ways.
        In this discussion, we assume a default representation which is a linear combination of a finite set of basis functions $\psi_{k}$:
        \begin{equation}\label{eq:rep:sum}
        \mathcal{V}({\bm r}) = \sum_{k} a_{k} \psi_{k}({\bm r}) .
        \end{equation}        
        Such representations (often accompanied by some penalties on large coefficients $a_k$) imply regularity of the objects; the specific type of regularity is determined by the choice of basis functions. 
        Typical examples of such functions would be low-frequency (band-limited) sine and cosine functions, and low-order polynomials. 
        The key properties of these representations are that once the model is formulated (i.e., once the basis functions are chosen), 
        the function $\mathcal{V}$ is completely determined by the coefficients $a_{k}$, and that the choice of basis functions imposes constraints or regularizes the function (a sum of low-frequency sines cannot yield a higher-frequency sine).
        
        In cryo-EM, the functions are sometimes described, loosely speaking, as ``band-limited'' and ``compactly supported.'' Often, these functions are defined through samples on 3-D grid, with different interpolations in different implementations. We represent functions with these properties in this work using generalized prolate spheroidal functions (see \cite{lederman2017prolate} and Section \ref{sec:res}), 
        however, the particular choice of basis functions is not the main topic of this paper, which applies to various forms of representation.

        A linear combination of basis functions is not the only way to represent functions. In particular, a Gaussian mixture model (GMM) has been proposed in \cite{kawabata2008multiple} for low-resolution representation of molecules in cryo-EM; in this representation, the function is a sum of Gaussian masses. 
        In this case, the coefficients determine the amplitude, centers and covariances of the masses. 
        The discussion in this paper also applies to representations like these, with some modifications. 
        In Sections \ref{sec:compositehyperobj} and \ref{sec:blackboxhyperobj} we extend the discussion to more general forms.

\begin{remark}[Terminology: ``representation'']
Our use of the term ``representation'' in the context of this paper is different from the
context in which we use the term in \cite{lederman2019representation}.
However, we have not found a better term that would avoid this confusion.
In this paper ``representation'' is a way of expressing a function or a problem,
typically an expansion of a function in some basis,
whereas in \cite{lederman2019representation} it is a technical representation theory term. 
These two works are independent; the conceptual relation between
the two is the motivation to treat heterogeneity as ``just another variable,''  analogous to the viewing direction variable.
\end{remark}

\subsection{Cryo-EM and the Forward Model}\label{sec:pre:cryo}

        The purpose of this section is to formulate the standard cryo-EM problem in the homogeneous case.
        We review the main characteristics of the cryo-EM imaging process and the forward model briefly,
        and discuss the Bayesian formulation of the problem of recovering the structure of a macromolecule.
        One of the ideas in this paper is to introduce a flexible framework where components can be exchanged for others to reflect slightly different models, therefore, we restrict the discussion in this section to the general structure of the formulation and highlight the key difficulties. 
        While it is certainly tempting to delve into the mathematical and numerical properties of the forward operator and the different parameters associated with it, the finer details are beyond the scope of this section. 
        A broader discussion of the imaging model and challenges can be found in many surveys such as \cite{frank,sigworth-review,cheng2015primer,milne2013cryo,vinothkumar2016single},
        and further discussions of a Bayesian framework for cryo-EM --- in the context of a maximum a posteriori (MAP) formulation ---
        can be found in \cite{scheres,scheres-relion}.
        We diverge slightly from the standard numerical representation of the homogeneous case in our use of generalized prolate spheroidal functions as
        natural basis functions for the problem (see Section \ref{sec:res}), but otherwise make use of a standard imaging model.

\begin{figure}
          \begin{center}
            \includegraphics[width=1.2in,trim={0.3in 0.15in 0 0},clip]{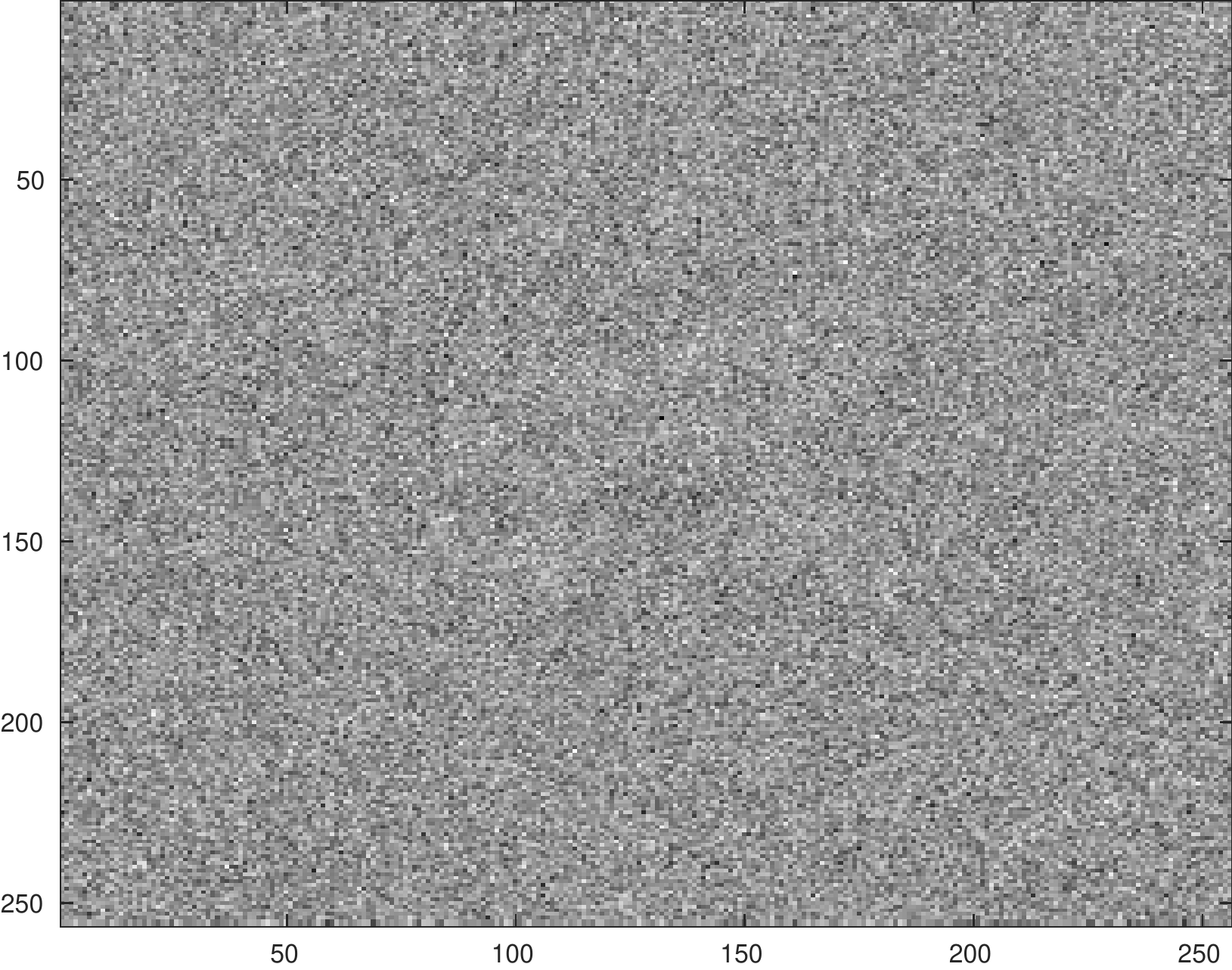}%
            ~\includegraphics[width=1.2in,trim={0.3in 0.15in 0 0},clip]{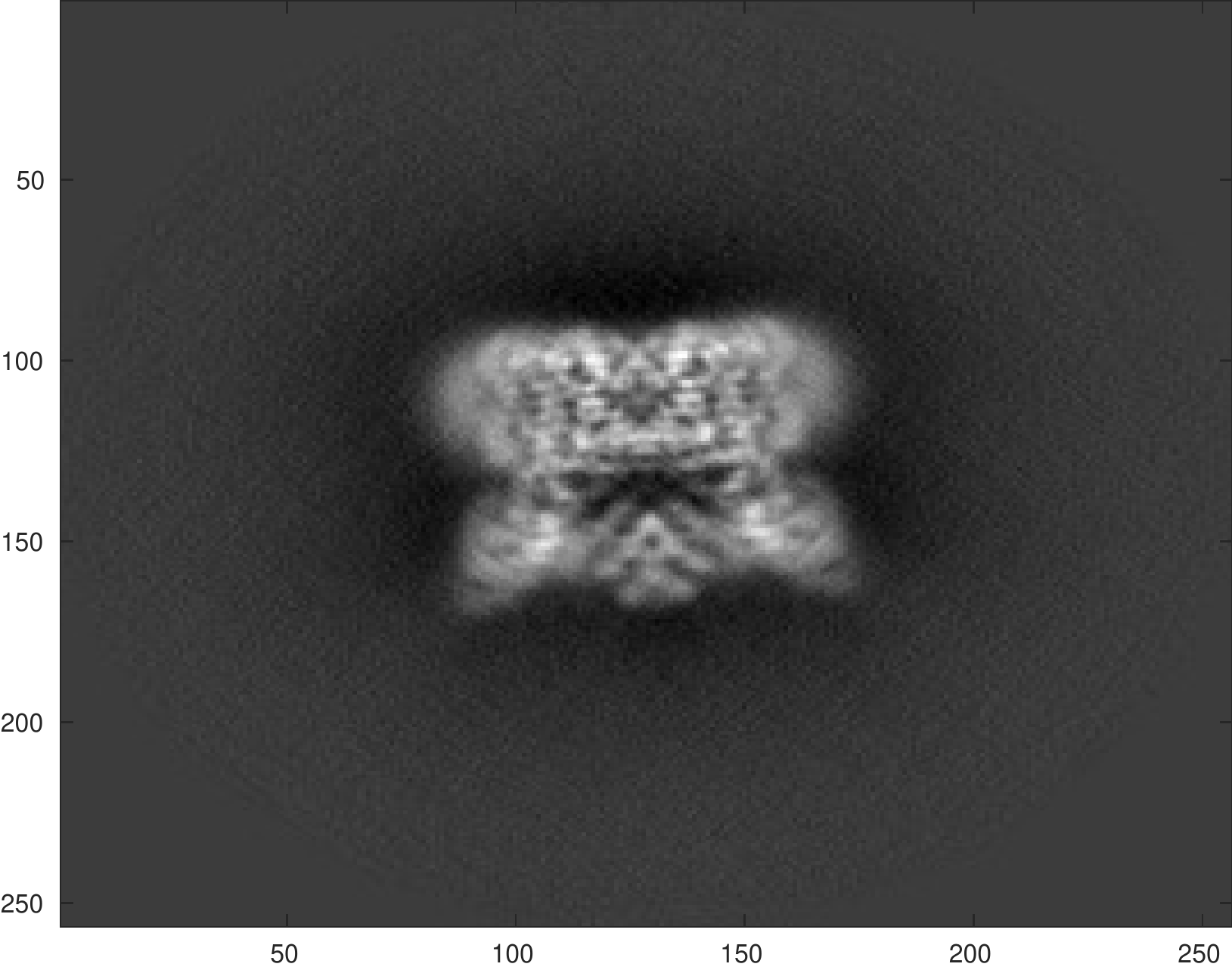}%

            \includegraphics[width=1.2in,trim={0.3in 0.15in 0 0},clip]{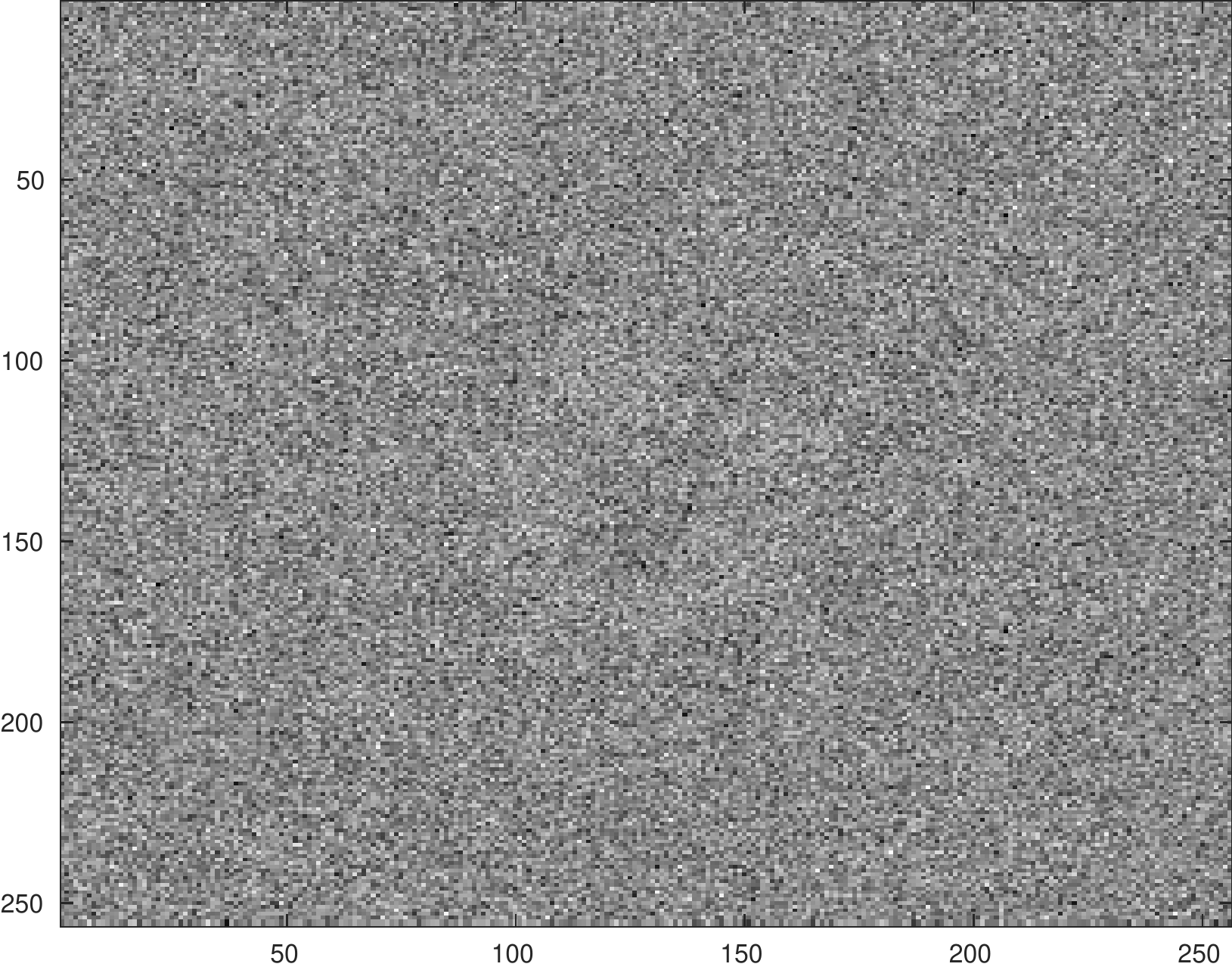}%
            ~\includegraphics[width=1.2in,trim={0.3in 0.15in 0 0},clip]{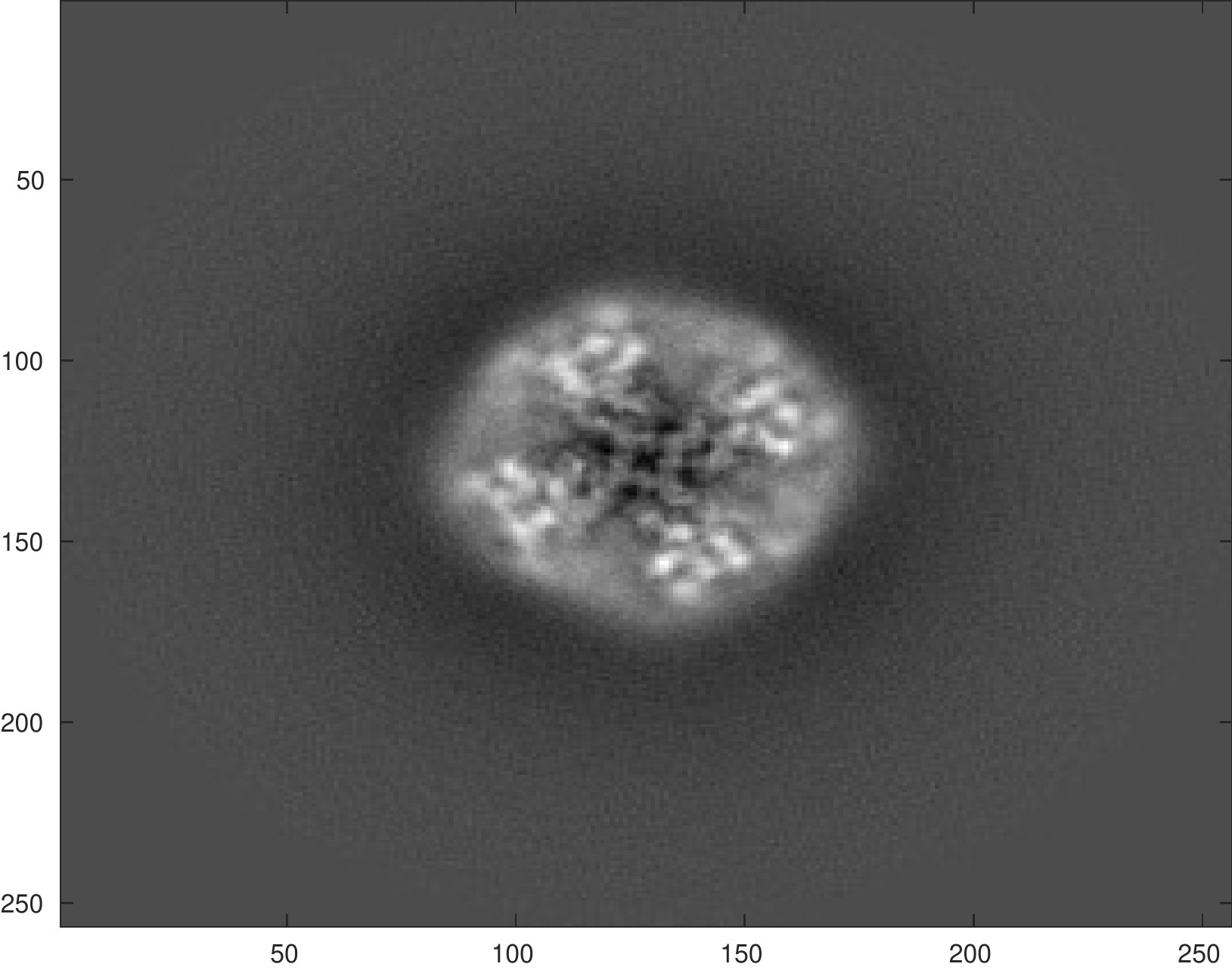}%
          \end{center}
          \caption{Left: two raw experimental images of  TRPV1, available via EMDB 5778 \cite{cheng-trpv1}. Right: computed projections of TRPV1 which are the close to the particle images on their left. 
          }\label{fig:cryosample}
        \end{figure}

        Electron microscopy is an important tool for recovering the 3-D structure of molecules.
        Of particular interest in the context of this paper is single particle reconstruction (SPR), and, more specifically, cryo-EM,
        where multiple 2-D projections, ideally of identical particles viewed from different directions,
        are used in order to recover the 3-D structure. Compared to other imaging problems, the cryo-EM  inverse problem is characterized by low SNR (illustrated in the example in Figure \ref{fig:cryosample}) and the unknown orientation of each particle image.

        The following formula is a simplified noiseless imaging model of SPR for obtaining
        the noiseless particle image $I^{(i)}$ from a function $\mathcal{V}$ (representing the molecule's  density or a potential):
	    \begin{equation} \label{eq:cryoemmodel:space}
		    I^{(i)}(r_x, r_y) = a_i \int H_i(r_x-r'_x , r_y - r'_y )   \left( \int_{\mathbb{R}} \mathcal{V}(R_i^{-1} {\bm {r'} + \bm {s}_i} ) d r'_z \right) d r'_x d r'_y,
	    \end{equation}
	    where  ${\bm r'}=(r'_x,r'_y,r'_z)^\intercal$, $H_i$ is a 2-D contrast transfer function (CTF) 
	    convolved with each 2-D projection of a particle, 
	    $R_i$ is the rotation that determines the direction from which the molecule is viewed,
	    ${\bm s}_i$ is the in-plane shift, and $a_i$ is a positive real valued contrast (amplitude).
	    The viewing direction $R_i$ and the in-plane shift $s_i$ are typically unknown. 
	    The parameters of the CTF are not all known; for simplicity, we will assume in this simplified model that they are known or estimated by other means. 
	    
	    A Fourier transform of both sides of Equation (\ref{eq:cryoemmodel:space}) reveals that, 
	    in the Fourier domain, the Fourier transform of the image $\hat{I}^{(i)}$ is related to the 3-D Fourier transform $\hat{\mathcal{V}}$ of the density ${\mathcal{V}}$ by the formula
        \begin{equation} \label{eq:cryoemmodel:freq}
		    \hat{I}^{(i)}(\omega_1,\omega_2) = a_i \hat{H}_i(\omega_1,\omega_2) S[\mathbf{s}_i](\omega_1,\omega_2)  \hat{\mathcal{V}}(R_i^{-1} {\mathbf {\bm{\omega}} } ) ,
	    \end{equation}
	    where ${\bm \omega} = (\omega_1,\omega_2, 0)^\intercal$, $S[\mathbf{s}_i]$ is the shift operator in the Fourier domain (which is a pointwise multiplication in the Fourier domain), and $\hat{H}_i$ is the Fourier transform of the CTF. In other words, in the Fourier domain, this imaging model reduces to an evaluation of the Fourier transform  $\hat{\mathcal{V}}$ in the plane perpendicular to the viewing direction, and to pointwise multiplications to compute the effects of CTF, shift and contrast. 

        In practice, the particle image ${Y}^{(i)}$ obtained in experiments is discrete (composed of pixels) and noisy. 
        We will study ${Y}^{(i)}$ through its {\em discrete} Fourier transform (as implemented by the FFT) $\hat{Y}^{(i)}$ of ${Y}^{(i)}$,
        evaluated at regular grid points $\{(\omega_1(k),\omega_2(k))\}$ in the Fourier domain.
        First, with a minor abuse of notation, we define the discrete noiseless particle image  $\hat{I}^{(i)}[\cdot]$ by sampling $\hat{I}^{(i)}(\cdot)$ at the points $\{(\omega_1(k),\omega_2(k))\}$ in the Fourier domain:
        \begin{equation} \label{eq:cryoemmodel:discrete}
		    \hat{I}^{(i)}[k] =  \hat{I}^{(i)}(\omega_1(k),\omega_2(k)) .
	    \end{equation}
	    We note that $\hat{I}^{(i)}(\omega_1(k),\omega_2(k)) =  \overline{\hat{I}^{(i)}(-\omega_1(k),-\omega_2(k))} $ and $\hat{I}^{(i)}(0,0)$ is real-valued, because ${I}^{(i)}$ is real-valued by definition. 
	    
	    For brevity and generality, we absorb the various imaging parameters such as the in-plane shift $s_i$ and contrast $a_i$ (as well as noise and CTF variables where applicable) of each particle image into an imaging variable which we denote by ${\bm{q}}_i$.	    
	    For the purpose of this discussion, we denote the forward model operator by $A( R_i, {\bm q}_i)$. The noiseless imaging model is then summarized by the formula
	    \begin{equation}
            I^{(i)} = A( R_i, {\bm q}_i) \mathcal{V}.
        \end{equation}
	    The map $A( R_i, {\bm q}_i)$ is typically linear.

        Next, we model the noise in a simplified imaging model for $\hat{Y}^{(i)}$:
        \begin{equation} \label{eq:cryoemmodel:noise}
		    \hat{Y}^{(i)}[k] =  \hat{I}^{(i)}[k] + {\sigma_k} \eta_{i,k} = \left( A( R_i, {\bm q}_i) \mathcal{V} \right)[k] + {\sigma_k} \eta_{i,k},
	    \end{equation}
	    where $\mathrm{Re}(\eta_{i,k})\sim N(0,1/2)$ and $\mathrm{Im}(\eta_{i,k})\sim N(0,1/2)$ are i.i.d, except for $\eta_{i,k} = \overline{\eta_{i,k'}} $ if $(\omega_1(k),\omega_2(k)) = (-\omega_1(k'),-\omega_2(k'))$ 
	    since the noisy image is real valued in the spatial domain. The sample at $\omega = 0$ has no imaginary component for the same reason. 
	    The noise variance ${\sigma_k}$ depends on the frequency; in this simplified model, we assume that the noise variance is known and is similar for all particle images; in practice it can be one of the model variables. 
	    
	    These simplified models neglect several aspects of the physical model, numerical computation, and experimental setup. For example, in practice, the images of individual particles must first be extracted from a larger image (micrograph).
	    As we noted above, the parameters determining the CTF and noise profile are sometimes added to the model. 
	    To allow a more general formulation, we add the variable ${\bm \mu}$ which encodes latent variable of the experiment that are not particle-specific (e.g., the noise standard deviation $\sigma_k$).
	    
	    Given this model, the likelihood $P(Y^{(i)} |  R_i, {\bm q}_i, \mathcal{V})$ of a particle image $Y^{(i)}$ given the object $\mathcal{V}$ and particle-specific variables $R_i$ and ${\bm q}_i$ is given by
	    \begin{equation}\label{eq:image_liklihood}  
          P(Y^{(i)} |  R_i, {\bm q}_i, {\bm \mu}, \mathcal{V}) \propto \mathrm{exp} \left(
             \sum_{k} \frac{\left| \widehat{Y^{(i)}}[k]  - \reallywidehat{\left(A( R_i, {\bm q}_i) \mathcal{V}\right)}[k] \right|^2}{2 \sigma_{i,k}^2}  \right).
        \end{equation}

        This leads to a Bayesian description of the problem, with a probability density for an object, image parameters and observed images given by:
        \begin{equation}
            P\left( \{ Y^{(i)},R_i, {\bm q}_i \}_i , {\bm \mu}, \mathcal{V} \right) = P\left( \{ R_i, {\bm q}_i \}_i , {\bm \mu}, \mathcal{V} \right) \prod_i P(Y^{(i)} |  R_i, {\bm q}_i, {\bm \mu}, \mathcal{V}),
        \end{equation}
        where $ P\left( \{ R_i, {\bm q}_i \}_i , \mathcal{V} \right)$ is a prior for the molecule and the particle-specific variables such as the viewing direction.
        The posterior distribution of the variables given the data is therefore proportional to the right-hand side of this equation:
        \begin{equation}\label{equ:pre:cryo:bayesian:general}
            \hspace{-2em}P\left( \{ R_i, {\bm q}_i \}_i , {\bm \mu}, \mathcal{V} | \{ Y^{(i)} \}_i \right) \propto P\left( \{ R_i, {\bm q}_i \}_i , {\bm \mu}, \mathcal{V} \right) \prod_i P(Y^{(i)} |  R_i, {\bm q}_i, {\bm \mu}, \mathcal{V}).
        \end{equation}
        The variables $ \{ R_i, {\bm q}_i \}_i $ are particle image specific latent variables, while the object itself, represented by $\mathcal{V}$, is the variable of interest. 
        In other words, the distribution that we are interested in is 
        \begin{equation}
             \hspace{-2em}P\left(  \mathcal{V} | \{Y^{(i)}\} \right) = \int P\left( \{ R_i, {\bm q}_i \}_i , {\bm \mu}, \mathcal{V} | \{Y^{(i)}\} \right)  dR_1 dR_2... dR_n d{\bm q}_1 d{\bm q}_2 ... d{\bm q}_n d{\bm \mu}
        \end{equation}

        Often, we would use a simpler model which assumes a uniform prior for the viewing directions and independent particle-specific variables $ \{ R_i, {\bm q}_i \}_i $; we obtain the posterior
        \begin{equation}\label{equ:pre:cryo:bayesian:simple}
            P\left( \{ R_i, {\bm q}_i \}_i , \mathcal{V} | \{Y^{(i)}\} \right) \propto P( \mathcal{V} ) P(  {\bm \mu} ) \prod_i  P(Y^{(i)} |  R_i, {\bm q}_i,  {\bm \mu} \mathcal{V}) P( {\bm q}_i ),
        \end{equation}
        where $P( \mathcal{V} )$ is a prior for molecules (e.g., weighted norms of coefficients representing the molecule), and $P({\bm q}_i)$ is a prior for the random variables controlling each individual images, such as in-plane shifts.
        
        While this general framework is sufficient for the purpose of this paper, we note that in the very influential work of \cite{scheres,scheres-relion}, 
        a Bayesian framework was used to formulate the problem of recovering a molecule $\mathcal{V}$ as a MAP estimation problem,
        implemented using an expectation-maximization algorithm. 
        We choose a slightly different formulation and different algorithms for our purpose due to several technical and computational considerations discussed below.
        Different algorithms use slightly different models and may absorb different components of the model into different latent variables.

\subsection{Heterogeneity in Cryo-EM}\label{sec:pre:cryo:het}

    The description of the cryo-EM problem in Section \ref{sec:pre:cryo} assumes that all the particles in all the projection images are identical (but viewed from different directions).
    However, the particles in a sample are often not identical. 
    In some cases, several different types of macromolecules or different conformations of the same macromolecule are mixed together,
    and sometimes there is  some flexibility in the structure of the macromolecule, which is manifested as a continuum of slightly different versions of the molecule.
    The first case of distinct classes of macromolecules is called {\em discrete heterogeneity} and the second case is called {\em continuous heterogeneity}.
    In this paper we focus on continuous heterogeneity, although much of the discussion applies to discrete heterogeneity with small modifications.

    A primary goal of this paper is to generalize the mathematical formulation in Section \ref{sec:pre:cryo} to the heterogeneous case.

\subsection{Existing Methods in Cryo-EM and Related Work}\label{sec:pre:cryo:existing}

Many of the existing algorithms for cryo-EM try to estimate the maximum-likelihood or the MAP molecule $\mathcal{V}$ from models formulated roughly like the model in Section \ref{sec:pre:cryo} (see, for example, \cite{fred-ml,sigworth2010chapter,scheres}).
One of the popular methods for this is a family of expectation-maximization  
algorithms, implemented in software such as RELION \cite{scheres-relion,kimanius,zivanov2018new}. Another is based in part on stochastic gradient descent (SGD), implemented in cryoSPARC \cite{brubaker}. 
These algorithms alternate between estimating the viewing direction (or conditional distribution of viewing directions) for each particle image given the current estimate of the macromolecule and updating the estimate of the macromolecule given the estimated viewing direction for each particle image (or its distribution).
In these updates, the algorithm must compare each particle image to the estimated macromolecule as viewed from each (discretized) viewing direction, at each value of the other variables (most notably, the in-plane shifts). Naturally, this comparison is expensive.
In recent years, several algorithms have been very successful in solving the homogeneous case (no heterogeneity). Clever algorithms and heuristics which reduce the number of comparisons significantly, and efficient use of hardware components such as GPUs have made the recent implementation of these algorithms rather fast \cite{scheres-relion,kimanius,punjani2016building,brubaker}. 
Other approaches to the cryo-EM problem rely on similarity measure between images to align the images before estimating the structure of the molecule \cite{shatsky,singer2010detecting,shkolnisky2012viewing,bandeira2015non}.
{An MCMC algorithm, using Gibbs Sampling, has been proposed for coarse modeling in the homogeneous case using a Gaussian mixture model \cite{joubert2015bayesian}.}

In addition to homogeneous reconstruction, many of the methods mentioned above also accommodate discrete heterogeneity through a 3-D classification framework, where each particle image is assigned to a separate 3-D reconstruction by maximizing a similarity measure. Expectation-maximization algorithms, such as RELION \cite{scheres-relion}, generalize to discrete heterogeneity by estimating conditional joint distributions of orientations and discrete class assignment.
While this approach has led to impressive results, it requires significant human intervention in a process of successive refinement of the datasets to achieve a more homogeneous sample, and structures that are not well represented in the data tend to be lost \cite{sigworth-review}.

A few approaches have emerged to treat the continuous heterogeneity problem. The remainder of this section briefly surveys some of the main approaches that are guided directly by cryo-EM images; a broader discussion is available in the recent survey \cite{sorzano2019survey}.
The methods proposed in \cite{dashti,schwander2014conformations,frank2016continuous} first groups images by viewing direction then attempts to learn the manifold formed by the set of images for each of those directions.
Following this, the various direction-specific manifolds are registered with one another, and a global manifold is obtained.
A 3-D model may then be constructed for each point on that manifold, providing the user with a description of the continuous varying structure. 
This method requires a consistent assignment of viewing directions across all states, and relies on a delicate metric for comparing noisy images to which different filters have been applied. 
The method assumes that certain properties of the manifold are conserved across the different viewing directions and requires a successful and globally consistent registration of the manifolds observed in different directions, which is not always possible. 
Furthermore, complex heterogeneity with more degrees of freedom results in manifolds that are intrinsically high-dimensional; such high-dimensional manifolds are difficult to estimate without exponential increase in the number of samples, and become more difficult to align.
This method has been demonstrated in the mapping of the continuous heterogeneity of the ribosome.

More recently, the RELION framework has been extended to include multi-body refinement \cite{nakane2018characterisation}  
(also see \cite{amunts2014structure,wong2014cryo,zhou2015cryo,zhou2015cryo,bai2015sampling,ilca2015localized}).
In this approach, the user selects different rigid 3-D models that are to be refined separately from the main, or consensus, model.
Each separate sub-model is then refined separately, with its own viewing direction and translation,  allowing it to move with respect to the consensus model in a rigid-body fashion.
This method is limited to rigid-body variability in a few sub-volumes, and cannot handle non-rigid deformations or other types of variability.
In particular, the structure found at the interface between the sub-models is likely to vary as their relative positions vary, and it is therefore lost in this method.

The covariance estimation approach proposed in \cite{anden2018structural} does not rely on a particular model for heterogeneity, be it discrete or continuous.
Indeed, the authors present a method for characterizing continuous variability in synthetic data.
However, the covariance approach is adapted to a linear model of variability and is therefore not well-suited for continuous, and necessarily non-linear, variability.
Furthermore, the limited resolution of the reconstruction precludes the study of heterogeneity at higher level of detail.
Another approach has been to study the normal modes of perturbation of a macromolecular structure \cite{tama2002exploring,jin2014iterative}.

\subsection{Markov Chain Monte Carlo (MCMC)}\label{sec:pre:mcmc}

MCMC is a collection of methods which have been used in statistical computing for decades. 
The full extent of these methods is  beyond the scope of this paper. The purpose of this section is to briefly mention a few properties of some MCMC methods that will be useful in our discussion, while inevitably omitting some technical details. 
A review of MCMC can be found in many textbooks, such as \cite{brooks2011handbook}.

MCMC algorithms are designed to sample from a probability distribution by constructing a Markov chain (i.e., a model of transitions between states at certain probabilities), 
such that the desired distribution is the equilibrium distribution of the Markov chain. 
Often, like in this paper, the desired probability from which we wish to sample is the posterior distribution $P({\bm X}|{\bm Y})$ of a variable ${\bm X}$, given a statistical model and data ${\bm Y}$. 
Very often, we have access only to an unnormalized density $h({\bm X}|{\bm Y}) \propto P({\bm X}|{\bm Y})$, so that we can compute the ratio $h({\bm X}|{\bm Y}) / h({ \tilde{\bm X}}|{\bm Y})$ between densities at two states 
${\bm {X}}$ and ${ \tilde{\bm X}}$, but not $P({\bm X}|{\bm Y})$ and $P({ \tilde{\bm X}}|{\bm Y})$ directly.

The {\em Metropolis-Hastings (MH)} algorithm, which is the basis for many MCMC algorithms, is based on the following Metropolis-Hastings Update:
\begin{itemize}
    \item Given the state ${\bm X}^{(n)}$ at step $n$, propose a new state ${\tilde{\bm X}^{(n+1)}}$ with conditional probability given the current state ${\bm X}^{(n)}$. 
    The probability of proposing ${\tilde{\bm X}^{(n+1)}}$ given the current state ${\bm X}^{(n)}$ is denoted by $q({\bm X}^{(n)}, {\tilde{\bm X}^{(n+1)}} |{\bm Y})$. 
    MH can be implemented in different ways, with different methods for proposing a new state, each method has a different function $q$ associated with it. 
    \item Compute the {\em Hastings ratio:}
    \begin{equation}
        r({\bm {X}^{(n)}}, {\tilde{\bm X}^{(n+1)}}) = \frac{h({ \tilde{\bm X}^{(n+1)}}|{\bm Y})q({\tilde{\bm X}^{(n+1)}},{\bm {X}^{(n)}}|{\bm Y})}  {h({ {\bm X}^{(n)}}|{\bm Y})q({{\bm X}^{(n)}},{\tilde{\bm X}^{(n+1)}}|{\bm Y})} .
    \end{equation}
    \item Approve the transition to the new state (i.e., ${\bm {X}^{(n+1)}} = {\tilde{\bm X}^{(n+1)}}$) with probability 
    \begin{equation}
        a({\bm {X}^{(n)}}, {\tilde{\bm X}^{(n+1)}}) = \min( 1, r({\bm {X}^{(n)}}, {\tilde{\bm X}^{(n+1)}}) ).
    \end{equation}
    If the proposed state is rejected, the previous state is retained with ${\bm {X}^{(n+1)}} = {\bm {X}^{(n)}}$.
\end{itemize}

Over time, under some conditions, MCMC samples states ${\bm {X}^{(n)}}$ from the equilibrium distribution, which is designed in MH to be $P({\bm X}|{\bm Y})$.

\begin{remark}
The Metropolis algorithm is a special case of the Metropolis-Hastings algorithm, with the transition probability chosen such that $q({\bm X}, { \tilde{\bm X}} ) = q({ \tilde{\bm X}}, {\bm X})$.
\end{remark}

\begin{remark}\label{remark:mcmc:mixing} 
MCMC allows a composition of update rules in different steps. 
For example, at each step, a subset of variables can be updated separately given the other variables. 
\end{remark}

\begin{remark}
\label{remark:mcmc:gibbs}
Gibbs sampling is a version of MCMC where at each step the algorithm samples some of the variables conditioned on other variables.
It is used when the joint distribution of all the variables is difficult to compute, but it is computationally feasible to sample some of the variables at each step while holding other variables fixed.
Formally, this is a special case of MH. 
We mention this important variant here for completeness, but the algorithms described in this paper do not rely on this version of MCMC, which is often not trivial to compute for all variables.
\end{remark}

We reiterate that this brief discussion of MCMC is not a comprehensive overview. The purpose of this discussion is to emphasize that MCMC can, in principle, be used to sample from a complicated posterior distribution 
even when the normalization of this distribution is unknown, and that various update strategies can be mixed together in MCMC algorithms. 
Samples from the posterior produced by MCMC can be used to approximate an expected value of a variable, but also to study the uncertainty.

\subsection{Metropolis-Adjusted Langevin Algorithm (MALA)}\label{sec:pre:mala}

MALA is a MH algorithm where the update proposal is given by the formula

\begin{equation}\label{eq:langevin:2}
    {\tilde{\bm X}^{(n+1)}} = {{\bm X}^{(n)}} + \frac{\sigma^2}{2} \nabla \log P({\bm X}^{(n)}  |{\bm Y} )  +  \sigma {\tilde{\bm W}^{(n+1)}},
\end{equation}
where
\begin{equation}
     {\tilde{\bm W}^{(n+1)}} \sim N \left( {\bm 0} ,  I_d \right) .
\end{equation}
Here, $\nabla \log P({\bm X}^{(n)}  |{\bm Y} )$ is the gradient of the log-likelihood with respect to the variables. Note that the unnormalized  $h({\bm X}|{\bm Y})$ is sufficient for computing the MALA steps.
The parameter $\sigma$ is set by the user.

A positive definite preconditioner matrix $A$ can be added without changing the equilibrium distribution:
\begin{equation}
    {\tilde{\bm X}^{(n+1)}} = {{\bm X}^{(n)}} + \frac{\sigma^2}{2} A \nabla \log P({\bm X}^{(n)} | {\bm Y} )  +  \sigma \sqrt{A} {\tilde{\bm W}^{(n+1)}}.
\end{equation}

MALA is just an update rule for which the Hastings ratio can be computed as usual, making it a standard Metropolis Hastings update. 
The MALA algorithm is motivated by the Langevin stochastic differential equation.
Loosely speaking, the Langevin stochastic differential describes a stochastic process which is analogous to Equation (\ref{eq:langevin:2}),
with infinitesimally small updates (small $\sigma$); the equilibrium distribution of this stochastic process is $P(\tilde{\bm X}  |{\bm Y} )$.

Works such as \cite{welling2011bayesian} find relations between the Langevin equation and SGD, a key algorithm in the area of deep learning, which has also been applied to cryo-EM by cryoSPARC \cite{brubaker}.

{
\subsection{Hamiltonian Monte Carlo (HMC)}\label{sec:pre:hamiltonian}

Hamiltonian Monte Carlo (HMC) is a another MCMC algorithm, which does not use the MH propose-accept-reject algorithm. 
HMC does not require sampling from a conditional distribution (required in Gibbs updates), but rather uses the gradient of the log-likelihood (like MALA)
for a combination of deterministic steps (unlike MALA) and randomized steps. 
Due to the limited scope of this paper, and the complexity of ideas behind HMC, we refer the reader to one of the many resources about MCMC and HMC, such as \cite{brooks2011handbook}, for additional information.
In the context of this discussion, the key property of HMC is its use of the gradient, which we discuss in the context of MALA;
however, HMC often has more advantageous mixing properties compared to MALA.

}

\section{Hyper-Molecules}\label{sec:analysis}

\subsection{Toy Examples}\label{sec:toy}

The purpose of this section is to introduce synthetic examples which we will use to illustrate some of the ideas
and in numerical experiments. 

\subsubsection{The ``Cat'':}\label{sec:toy:cat}

\begin{figure}[h]
\begin{center}
  \includegraphics[width=0.45 \linewidth]{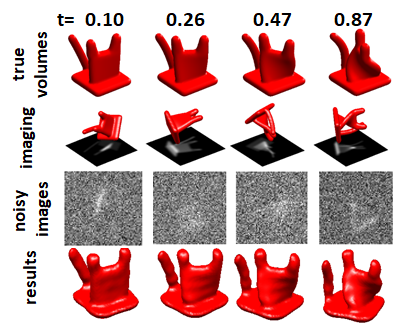}%
\end{center}
\caption{Sample cats: true 3-D instances (top row), rotated instance and noiseless projection images (second row), images with noise as used in the simulation (third row), and the reconstructed cat (bottom row,  discussed in more detail in an earlier technical report \cite{lederman2017continuously})}\label{fig:toy:cat}
\end{figure}

To illustrate the problem, we constructed the ``cat,''
an object composed of Gaussian elements
in real space, where each Gaussian follows a continuous trajectory as a function of the parameter $t$, so that 
we have a continuous space of objects corresponding to an object with extensive large-scale heterogeneity.
The heterogeneity is one-dimensional, where the state corresponds to the direction in which the cat's ``head'' is turned.
Examples of synthetic 3-D object instances and the 2-D projections are presented in Figure \ref{fig:toy:cat} (rows 1-3).

\subsubsection{The ``Pretzel'':}\label{sec:toy:Pretzel}

\begin{figure}[h]
\centering
\begin{center}
\includegraphics[width=0.3 \linewidth]{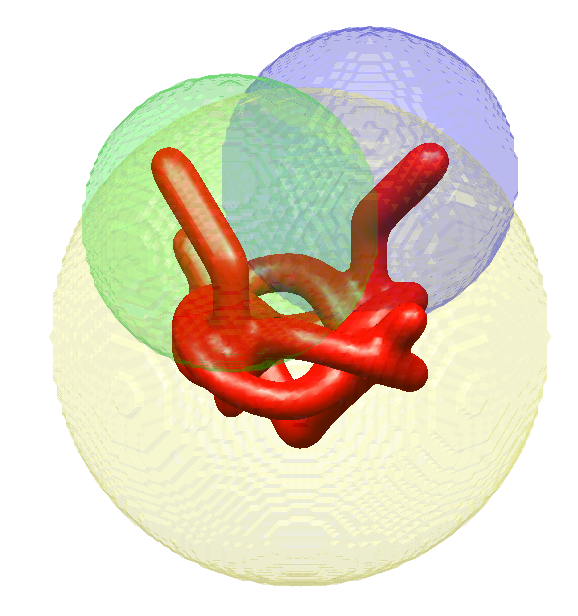}%
\caption{The anatomy of the pretzel: the green and blue regions identify the heterogeneous ``arms.'' 
In the analysis in Section \ref{sec:res}, the yellow region marked the boundary of the rigid component, and the green and blue balls marked the boundaries of the two heterogeneous components.} \label{fig:toy:pretzel:parts}
\end{center}
\end{figure}

\begin{figure}[h]
\begin{center}
\includegraphics[width=0.6 \linewidth]{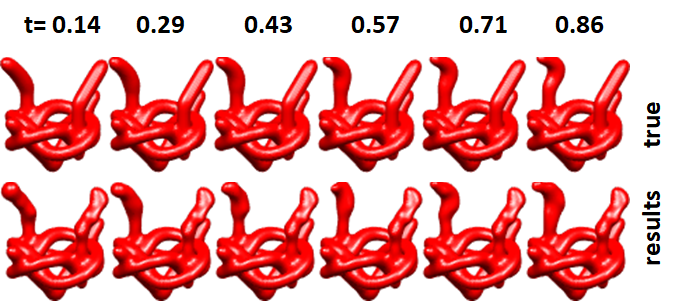}%
\caption{The pretzel: samples of true pretzels and reconstructed pretzels (see Section \ref{sec:res}). For the purpose of this illustration only, we hold one of the arms in a fixed state. In the simulation and the recovered object, the arms move independently. } \label{fig:toy:pretzel:exp}
\end{center}
\end{figure}

To illustrate continuous heterogeneity with more structure, we constructed the ``pretzel,''
which is composed of three parts: a rigid ``base'' and two independent ``arms.''
The two heterogeneous regions are highlighted in the green and blue balls in Figure  \ref{fig:toy:pretzel:parts}.
In Figure \ref{fig:toy:pretzel:exp}(top) we present different conformations of the pretzel. In our simulations, each arm can take any state independently of the other, but for the purpose of illustration in Figure \ref{fig:toy:pretzel:exp}, we hold one of the arms in a fixed state and sample different states of the other arm. 

This is a simplified illustrative mock-up of a typical experiment where one part of the macromolecule is rigid and others are heterogeneous and deforming. 
A dataset and a simulation using this model are described in Section \ref{sec:res}.

~

\subsection{Generalizing Molecules: Hyper-Molecules}\label{sec:hyperobj}

Hyper-molecules generalize 3-D density functions $\mathcal{V}({\bm r})$ to higher-dimensional functions $\mathcal{V}({\bm r},{\bm \tau})$ with the new state variable ${\bm \tau}$.
For a fixed conformation or state ${\bm \tau}$, the 3-D density function  $\mathcal{V}(\cdot,{\bm \tau})$ represents the molecule at that given conformation. 

To illustrate the idea, we consider the cat example in Section \ref{sec:toy:cat}. 
A natural way to view this cat is to produce a 3-D movie of the cat, where we would see a different conformation of the cat in each frame of the movie.
In other words, each frame would present $\mathcal{V}(\cdot,{\bm \tau})$ for a different value of ${\bm \tau}$.
Since the deformation of the cat is continuous, we could sample it at any arbitrary value of ${\bm \tau}$; a viewer may expect the movie to show a continuous 
transformation, with the cat not changing considerably as we move from one frame to the next. In other words, the movie would be expected to be relatively smooth (with several possible definitions of smoothness). 
This property of the movie reflects relations between different conformations. Hyper-molecules enforce such relations in the modeling of $\mathcal{V}(\cdot,{\bm \tau})$.

We recall that density functions in cryo-EM are often assumed to be band-limited, effectively making them smooth in the spatial domain. 
This regularity is enforced by the representation defined in (\ref{eq:objgeneral})  where the basis functions $\psi_k$ are approximately band-limited.
Hyper-molecules enforce regularity in the state space through the definition in  (\ref{eq:hyperobjgeneral}) by choosing $\tilde{\psi}_k$ that have a similar regularity property in the state variable. 
For example, in the case of 1-D state space in the cat example, with the state variable representing the direction in which the cat is looking, 
a natural generalization of the representation in (\ref{eq:objgeneral}) generates 4-D basis functions $\tilde{\psi}_{k,q}({\bm r},t)$ 
from products $\tilde{\psi}_{k,q}({\bm r},t) = {\psi}_{k}({\bm r}) P_q(t)$ of 3-D functions ${\psi}_{k}$ and low-degree orthogonal polynomials $P_q$ (e.g., Chebyshev polynomials)
such that
\begin{equation}\label{eq:hyperobj:1D}
  \mathcal{V}({\bm r},\tau) = \sum_{k,q} a_{k,q} \psi_k({\bm r}) P_q(t).
\end{equation}

More generally, when there are $d$ degrees of freedom of flexible motion, the manifold of conformations is of dimension $d$ and the time variable $t$ in Equation (\ref{eq:hyperobj:1D}) is replaced by manifold coordinates ${\bm \tau} \in T$. The polynomials $P_q$ are replaced by a truncated set of basis functions over the manifold, denoted $P_q({\bm \tau})$, with a minor abuse of notation:
\begin{equation}\label{eq:hyperobj}
  \mathcal{V}({\bm r},{\bm \tau}) = \sum_{k,q} a_{k,q} \psi_k({\bm r}) P_q({\bm \tau}).
\end{equation}
For example, the basis function $P_q({\bm \tau})$ can be the product of polynomials in multiple variables.

The model in Section \ref{sec:pre:cryo} then generalizes naturally, such that Equation (\ref{eq:cryoemmodel:space}) is generalized to 
\begin{equation} \label{eq:cryoemmodel:het:space}
		    I^{(i)}(x_1,x_2) = a_i H_i * \int_{\mathbb{R}} \mathcal{V}(R_i^{-1} {\mathbf{r} + \mathbf{s}_i, {\bm \tau}_i} ) d x_3 ,
	    \end{equation}
the corresponding operator $A( R_i, {\bm q}_i)$ to $A( R_i, {\bm \tau}_i, {\bm q}_i)$, 
and the posterior (\ref{equ:pre:cryo:bayesian:general}) to 
\begin{equation}\label{equ:hyperobj:bayesian:general}
           \hspace{-2em} P\left( \{ R_i,  {\bm \tau}_i, {\bm q}_i \}_i , {\bm \mu}, \mathcal{V} | \{ Y^{(i)} \}_i \right) \propto P\left( \{ R_i,  {\bm \tau}_i, {\bm q}_i \}_i , {\bm \mu}, \mathcal{V} \right) \prod_i P(Y^{(i)} |  R_i,  {\bm \tau}_i, {\bm q}_i, {\bm \mu}, \mathcal{V}).
        \end{equation}

In other words, we use the formulation of the continuously heterogeneous molecules as hyper-molecules to  generalize the Bayesian formulation of the cryo-EM problem 
from a problem of recovering a 3-D molecule from 2-D projections in unknown viewing directions to a problem of recovering a higher dimensional hyper-molecule from 2-D projections. 
The key to this formulation, compared to a formulation as a collection of independent molecules (e.g., \cite{van2006four,scheres-relion}), is that hyper-molecules encode relations between states, 
with the related property that they encode a smoothly varying continuum of states.

\subsection{Enforcing Structure}\label{sec:rep}

We note that there exists an equivalent scheme using appropriate samples in the state space, which would be numerically equivalent to our use of polynomials in the state variable.
However, hyper-molecules are different from independent molecules because they provide relations between states. This regularity in the relation between states can be further reinforced by generalizing other ideas implemented for 3-D molecules such as priors that favor smaller coefficients for basis functions with high-frequency components in the state variable.
Furthermore, the interpolation allows us to assign to each particle image any state in the continuum, rather than only the sampled states.

The basis functions presented above are not the only way to define such relations between states; for example, one can use a discretized state space and use linear interpolation between sorted discretized states (equivalent to a basis of triangles in the state space) to obtain a continuum of states. 
One case also  penalize for large changes between adjacent states using a term of the form
\begin{equation}
 L(\mathcal{V}) = \sum_{t=1}^{T-1} \int_{} | \mathcal{V}({\bm r},t) - \mathcal{V}({\bm r},t+1) |^2 d {\bm r}.
\end{equation}
In fact, smoothness and continuity are crude proxies for properties that we would expect to find in the state space of molecules. 
For example, often, we would expect to  observe a flow of mass as we move between states. This would be captured better through 
a Wasserstein distance between states; additional physical properties are discussed in the remainder of the paper and in a technical report \cite{lederman2017continuously}.
In the Bayesian formulation, it is natural to add explicit priors for hyper-molecules.

\subsection{A Curse of Dimensionality}\label{sec:dim}

Building upon the success of the maximum likelihood and MAP frameworks in cryo-EM (see discussion in Section \ref{sec:pre:cryo:existing}), it is natural to consider their application to the hyper-volume reconstruction problem. 
The expectation-maximization algorithms are iterative refinement algorithms which attempt to recover the maximum-likelihood or MAP solution by alternating between updating the distributions of variables such as the viewing direction $R_i$ and updating the estimate of the molecule $\mathcal{V}$ (i.e.,  coefficients in the representation of the object as defined in Equation (\ref{eq:objgeneral})). 
Generating the projections for all viewing directions and comparing them to all particle images are computationally intensive operations in the implementation.

In the case of hyper-molecules, expectation-maximization would be generalized to alternating between updating the joint distribution over viewing directions $R_i$ and  (possibly high-dimensional) state variables ${\bm \tau_i}$ (compared to a small number of discrete conformations in current algorithms) and updating the hyper-molecule (\ref{eq:hyperobj}). In other words, one would have to project the hyper-molecule in every possible state in every possible viewing direction and compare each particle image with each of these projections, rapidly increasing the number of comparisons in this already expensive procedure. More complex models of hyper-molecules, introduced later in this paper, would make it more difficult to design specialized algorithms and heuristics to optimize this procedure.

In addition, we note that the number of coefficients required to represent a molecule as a linear combination of basis functions in Equation (\ref{eq:objgeneral}),
at a resolution corresponding to about $N\times N \times N$  voxels, is $O(N^3)$. 
Similarly, adding $d$-dimensional heterogeneity at ``state space resolution'' corresponding to $Q$ state coefficients requires $O(N^3 Q^d)$ coefficients.
High-dimensional heterogeneity, arising, for example, in molecules that have several independent heterogeneous regions, results in a large number coefficients which could exceed the total number of pixels in all particle images of an experiment. 
Indeed, since hyper-molecules have the capacity to represent very generic molecules, it is natural to expect that a lot of data would be required to estimate them;
in particular, if the number of possible states (in some discretization) grows exponentially fast with the dimension $d$, it is natural to expect the required number of particle images to grow as fast, if not faster. 
Given infinite data and infinite computational resources, it is tempting to model very little and allow the data and algorithm to reveal the structure.
Unfortunately, despite the rapid growth in cryo-EM throughput and computational resources, they are far from ``infinite.''
The natural question to ask is if we can use prior knowledge and assumptions to reduce the amount of data that we need, even in the case of high-dimensional heterogeneity.

In the remainder of this paper, we address some of these challenges.

\subsection{Finer Structures I: Composite Hyper-Molecules}\label{sec:compositehyperobj}

In the previous section, we found that recovering a hyper-molecule which describes very generic, and potentially complicated, dynamics of a macromolecule requires massive amounts of data.
Often, researchers  have prior knowledge about the structure and dynamics of the macromolecule that they study. 
For example, many macromolecules are composed of a static component to which smaller flexible heterogeneous components are attached (for an illustrative toy example, see the pretzel example in Section \ref{sec:toy:Pretzel}).
Often, practitioners are able to use traditional cryo-EM algorithms to recover the static component at high resolution, but the regions of the flexible components are blurry. 
In these cases, researchers are often able to hypothesize where each component is located, which components are static, and which components are heterogeneous. 
Tools for estimation of local variance and resolution help researchers in identifying these regions (see,  for example, \cite{LiuFrank1995,Penczek2002,PenczekEtal2006,LiaoFrank2010,PenczekKimmelSpahn2011,AndenKatsevichSinger2015,AndenSinger2018}).

We introduce {\em composite hyper-molecules}, a model which is the sum of $M$ components $\mathcal{V}^{m}$, each of which is a hyper-molecule. 
The following formula describes a simple version of a composite hyper-molecule:
\begin{equation}\label{eq:compositehyperobj:simple}
  \mathcal{V}({\bm r},{\bm \tau}^1, {\bm \tau}^2, ...., {\bm \tau}^M) =  \sum_{m=1}^M \mathcal{V}^{m} ( {\bm r} ,{\bm \tau}^m).
\end{equation}
Each component is constrained to a certain region of space where it is assumed to be supported (the regions may overlap). 
Each component has its own set of state variables and coefficients that describe it. 
In our pretzel example, the yellow region in Figure \ref{fig:toy:pretzel:parts} is modeled as a rigid static ``body,'' and the green and blue regions represent regions of space where two one-dimensional heterogeneous components are supported. As can be seen in this example, the regions may overlap and do not have to be tight around the actual component.

In some cases, the different components could be roughly described as moving one with respect to the other, in addition to more subtle deformations (for example, at the interface between the components).
Indeed, heterogeneous macromolecule have been modeled as a superposition of several rigid objects in somewhat arbitrary relative positions in work such as \cite{nakane2018characterisation,amunts2014structure,wong2014cryo,zhou2015cryo,zhou2015cryo,bai2015sampling,ilca2015localized}.
We observe that hyper-molecules and the composite hyper-molecules in Equation (\ref{eq:compositehyperobj:simple}) are generic enough to describe the relative motion of these components, but if such dynamics can be assumed,
capturing them in the model is advantageous for computational and statistical reasons.  
Therefore, a more complete version of composite hyper-objects allows both motion and heterogeneity in each component
\begin{equation}\label{eq:compositehyperobj:traj}
\eqalign{
  \mathcal{V}({\bm r},{\bm \tau}^{1,\mathrm{state}}, {\bm \tau}^{2,\mathrm{state}}, ...., {\bm \tau}^{M,\mathrm{state}}, {\bm \tau}^{1,\mathrm{position}}, {\bm \tau}^{2,\mathrm{position}}, ...., {\bm \tau}^{M,\mathrm{position}}) = \cr
  {\sum_{m=1}^M \mathcal{V}^{m} ( f^m ({\bm r},{\bm \tau}^{m,\mathrm{position}}),{\bm \tau}^{m,\mathrm{state}}) }
  }
\end{equation}
where $f^m ({\bm r},{\bm \tau}^{m,\mathrm{position}})$ is a function that describes the trajectory of the $m$th component, 
so that the component is in heterogeneity state ${\bm \tau}^{m,\mathrm{state}}$ 
and its location along the ``trajectory'' is determined by the position variable ${\bm \tau}^{m,\mathrm{position}}$.
For example, a simple affine $f^m$ can take the form 
\begin{equation}\label{eq:compositehyperobj:traj:example}
f^m({\bm r},{\bm \tau}^{m,\mathrm{position}} ) = \left( \eqalign{ {\tau}^{m,\mathrm{state}}{\theta}^{m,\mathrm{position}}_{x,1} + {\theta}^{m,\mathrm{position}}_{x,0}+ r_x \cr {\tau}^{m,\mathrm{state}}{\theta}^{m,\mathrm{position}}_{y,1} + {\theta}^{m,\mathrm{position}}_{y,0} +r_y \cr
{\tau}^{m,\mathrm{state}}_i{\theta}^{m,\mathrm{position}}_{z,1} + {\theta}^{m,\mathrm{position}}_{z,0} + r_z } \right), 
\end{equation}
where ${\bm r}=(r_x,r_y,r_z)^\intercal$. 
The variables ${\bm \theta}^{m,\mathrm{position}}$, which determine the trajectory, are part of the variables describing the hyper-molecule, much like the coefficients in Equation (\ref{eq:compositehyperobj:simple}).
Actual trajectory functions would presumably be more complex and could involve rotations and deformations.

The variables for the position ${\bm \tau}^{m,\mathrm{position}}$ and state ${\bm \tau}^{m,\mathrm{state}}$ can be closely related
(the position can be related to the heterogeneity state variable for that component);
for brevity, we use ${\bm \tau}^{m}$ as a state variable that encapsulates both ${\bm \tau}^{m,\mathrm{position}}$ and ${\bm \tau}^{m,\mathrm{state}}$. 

Compared to previous work like \cite{nakane2018characterisation,amunts2014structure,wong2014cryo,zhou2015cryo,zhou2015cryo,bai2015sampling,ilca2015localized}, the composite hyper-molecule formulation models components that are inherently non-rigid, and, in particular, models the flexible interface between components. 
Furthermore, composite hyper-molecules model the set of possible relative positions (trajectories) of the different components with respect to each other (as opposed to more arbitrary possible relative positions), which are parametrized and fitted using data.

\begin{remark}
    In some cases, there are relations between the different regions that can be captured in the description of the composite hyper-molecule. 
    For example, our pretzel has two identical arms (shifted and rotated with respect to each other). While each arm can appear in a different state independently from the other arm, they have the same fundamental structure (i.e., they are the same hyperobject, at a different state and position). A similar phenomenon is observed in some macromolecules that have certain symmetries.
    We capture this fact in our model by defining the hyper-objects representing the two arms so that they share coefficients in their representation. This is analogous to ``weight sharing'' in deep neural networks. 
\end{remark}

\subsection{Finer Structures II: Priors and ``Black-Box Hyper-Molecules'' }\label{sec:blackboxhyperobj}

The purpose of this section is to add a layer of abstraction to the modeling of hyper-molecules,
where the model can be implemented as a ``black box'' provided to an algorithm designed to recover hyper-molecules;
the algorithms themselves are discussed in later sections, while this section focuses on the formal modeling of these components.
These black-box models will allow users with different levels of technical expertise to define more elaborate models and priors which reflect assumptions and prior knowledge about the experiment,
to the extent that such assumptions are necessary given the amount of data, model complexity and available computational resources.
While the implementation presented in this paper treats simpler models, this section provides context for goals of this line of work, and additional motivation for algorithms guided by gradients (MALA and HMC) and for the work on MCMC algorithms.

We revisit the formulation of the hyper-molecule $\mathcal{V}$ as a sum of basis functions in Equation (\ref{eq:hyperobj}).
We denote the coefficients of these basis functions by ${\bm \theta}$. Similarly, in the formulation in Equation (\ref{eq:compositehyperobj:traj}),
the coefficients of the basis functions in all components and the coefficients of the trajectories are denoted collectively by ${\bm \theta}$.
We write this fact explicitly using the notation $\mathcal{V}[{\bm \theta}](r,{\bm \tau})$.
We revisit Equation (\ref{equ:hyperobj:bayesian:general}), and add this explicit notation:
\begin{equation}\label{equ:hyperobj:bayesian:blackbox1}
           \hspace{-6em} P\left( \{ R_i,  {\bm \tau}_i, {\bm q}_i \}_i , {\bm \mu}, \mathcal{V}[{\bm \theta}] | \{ Y^{(i)} \}_i \right) \propto P\left( \{ R_i,  {\bm \tau}_i, {\bm q}_i \}_i , {\bm \mu},{\bm \theta} \right) \prod_i P(Y^{(i)} |  R_i,  {\bm \tau}_i, {\bm q}_i, {\bm \mu}, \mathcal{V}[{\bm \theta}]).
        \end{equation}

In particular, it is compelling to factorize (\ref{equ:hyperobj:bayesian:blackbox1}) into simpler components and formulate a  more specific structure:
\begin{equation}\label{equ:hyperobj:bayesian:blackbox2}
\eqalign{
            P\left( \{ R_i,  {\bm \tau}_i, {\bm q}_i \}_i , {\bm \mu}, \mathcal{V}[{\bm \theta}] | \{ Y^{(i)} \}_i \right) \propto \cr 
            P\left( {\bm \theta} \right) 
            P\left(  {\bm \mu}\right)
            \prod_i P(Y^{(i)} |  R_i,  {\bm \tau}_i, {\bm q}_i, {\bm \mu}, \mathcal{V}[{\bm \theta}])  P\left(   R_i,  {\bm \tau}_i, {\bm q}_i  | {\bm \mu} \right).
            }
        \end{equation}
        where $ P\left( {\bm \theta} \right)$ is a black-box prior for the hyper-molecule, 
        $P\left(  {\bm \mu}\right)$ is a black-box prior for imaging variables and latent variables (e.g., noise parameters and CTF parameters for micrographs),
        $P\left(   R_i,  {\bm \tau}_i, {\bm q}_i  | {\bm \mu} \right)$ is a prior for the variables of each particle image (e.g., shift from center, contrast parameters), 
        and $P(Y^{(i)} |  R_i,  {\bm \tau}_i, {\bm q}_i, {\bm \mu}, \mathcal{V}[{\bm \theta}])$ is the relation to the measurements.

        In this formulation, $\mathcal{V}$ can be replaced by an arbitrary black-box function that produces a consistent notion of a hyper-molecule; 
        this black-box formulation decouples the specifics of the model from the algorithm, giving the scientist more flexibility in defining their model.
        The key components in this formulation are  the model $\mathcal{V}[{\bm \theta}]$ which defines the density at any spatial position and state as a function of the coefficients ${\bm \theta}$, 
        and a prior $P\left( {\bm \theta} \right)$. These two components encode the scientist's assumptions, prior knowledge and physical constraints.
        Another key component is $P(Y^{(i)} |  R_i,  {\bm \tau}_i, {\bm q}_i, {\bm \mu}, \mathcal{V}[{\bm \theta}])$, which encapsulates the imaging model. 
        The components $P\left(   R_i,  {\bm \tau}_i, {\bm q}_i  | {\bm \mu} \right)$ and $ P\left(   R_i,  {\bm \tau}_i, {\bm q}_i  | {\bm \mu} \right)$ give some additional flexibility in modeling.
        
       Having defined the models, we turn to the discussion of the algorithms. The general black-box form of the models presented in Section \ref{sec:blackboxhyperobj} above provides some of the motivation for algorithms that are compatible with such generic model.

\section{Algorithms}\label{sec:alg}

In this section we discuss the role of MCMC algorithms in the framework for recovering hyper-molecules.

\subsection{MCMC, MALA and HMC}

We consider the Bayesian formulation of  hyper-molecules in Equation (\ref{equ:hyperobj:bayesian:blackbox2}).
The difficulty with expectation-maximization algorithms is that they compute  $P( R_i,  {\bm \tau}_i, {\bm q}_i | Y^{(i)}, {\bm \mu}, \mathcal{V}[{\bm \theta}])$
as a function of all possible combinations of viewing directions $R_i$, states ${\bm \tau}_i$, and some of the other particle-image specific variable ${\bm q}_i$ (e.g., in-plane shift) at every iteration
(the update of ${\bm \theta}$ involves another computationally expensive operation for similar reasons).
This involves some discretization of these variables and a large number of comparisons which are computationally expensive at every iteration. 
This is a computational challenge in the homogeneous case and in the case of discrete heterogeneity when there is a small number of conformations;
the natural generalization to high-dimensional continuous heterogeneity increases the computational complexity exponentially in the dimensionality of the heterogeneity.
Indeed, algorithms and heuristics have been developed for reducing the number of comparisons in existing software, but it is a challenge to generalize them to apply to high-dimensional hyper-molecules and generic black-box models whose specific form is defined by a user and is not available when the software is written. 

We propose an MCMC framework for sampling from the posterior in Equation (\ref{equ:hyperobj:bayesian:blackbox2});
some of the main features of MCMC are reviewed briefly in Section \ref{sec:pre:mcmc}.
We note that MCMC is not a single algorithm, but a collection of algorithms that can be used together. 

Equation (\ref{equ:hyperobj:bayesian:blackbox2}) and the analogy to expectation-maximization suggests that different variables in the MCMC formulation can be treated separately, mixing strategies for updating a subset of variables while holding the others constant. 
In particular, the particle-image variables $R_i,  {\bm \tau}_i$ and $ {\bm q}_i $ can be evaluated separately and in parallel because they are independent conditioned on ${\bm \mu}$ and $\mathcal{V}[{\bm \theta}]$.
MCMC algorithms such as a simple MH (with a simple update strategy) do not require the computation of the distribution 
$P(Y^{(i)} |  R_i,  {\bm \tau}_i, {\bm q}_i, {\bm \mu}, \mathcal{V}[{\bm \theta}])$ for every value of 
$R_i,  {\bm \tau}_i$ and ${\bm q}_i$, but rather require only the ratio 
$P(R_i,  {\bm \tau}_i, {\bm q}_i | Y^{(i)} ,   {\bm \mu}, \mathcal{V}[{\bm \theta}]) / P( \tilde{R}_i,  { \tilde{\bm \tau}}_i, { \tilde{\bm q}}_i | Y^{(i)} , {\bm \mu}, \mathcal{V}[{\bm \theta}])$ between 
the likelihoods of different values of the variables;
in other words, at every iteration, this version of MCMC requires only the evaluation at two points in the update of particle-image specific variables
and it is sufficient to have $P(Y^{(i)} |  R_i,  {\bm \tau}_i, {\bm q}_i, {\bm \mu}, \mathcal{V}[{\bm \theta}])$ up to a multiplicative constant (so that the probability does not need to be normalized to integrate to $1$).
Other strategies, such as MALA and HMC, require the gradient of the log-likelihood with respect to the different variables (again, implying that the probability does not need to be normalized to integrate to $1$).
Similar considerations apply to the update of other variables. 
We note that MCMC is not a ``magic solution'' to the computational challenge, because it may require more steps than expectation-maximization, 
but each step is computationally tractable and different strategies and  tools can easily be combined to improve performance; where expectation-maximization is feasible, analogous MCMC steps can be applied.

MCMC yields a sample of the variables and latent variable; we can restrict our attention to variables such as ${\bm \theta}$ which are sampled hyper-molecules, 
and we can consider the statistics of ${\bm \tau}$ if we wish to study the statistics of states' occupancy.
Most often, in practice, ${\bm \theta}$ or  $\mathcal{V}$ can be averaged over all the samples to produce an ``expected'' hyper-molecule, although this averaging can introduce some technical difficulties due to ambiguities which we will discuss briefly later; these technical issues are not uncommon in this type of problems, and in practice they are rarely a problem.
A similar problem happens the the maximum-likelihood and MAP approaches, since there are several equivalent solutions. There too, this is not a problem in practice. 
The advantage of having multiple samples from the posterior, however, is that they allow us to study the uncertainty in the solution by studying the variability of $\mathcal{V}$.

\subsection{A Remark about Black-Box Hyper-Molecules}

In this section, we revisit the Bayesian formulation of Equation (\ref{equ:hyperobj:bayesian:blackbox2}) and discuss some aspects of the formulation of generalized hyper-molecules
that are related to the algorithms and implementation.
In principle, it is sufficient to define black-box functions which would evaluate the prior $P\left( {\bm \theta} \right)$ and the density $\mathcal{V}[{\bm \theta}]({\bm{r}},{\bm \tau})$ at any spatial (or frequency) location ${\bm{r}}$, and any state ${\bm \tau}$ (and possibly provide the interface for computing gradients over the difference variables); 
the algorithm would use these functions to compute $P(Y^{(i)} |  R_i,  {\bm \tau}_i, {\bm q}_i, {\bm \mu}, \mathcal{V}[{\bm \theta}])$ using its imaging model. 

We note that the explicit evaluation of $\mathcal{V}[{\bm \theta}]$ is not required in Equation (\ref{equ:hyperobj:bayesian:blackbox2}). Instead, $\mathcal{V}$ is considered implicitly in the prior $P\left( {\bm \theta} \right)$
and  in the comparisons to images in $P(Y^{(i)} |  R_i,  {\bm \tau}_i, {\bm q}_i, {\bm \mu}, \mathcal{V}[{\bm \theta}])$.
The way that $\mathcal{V}[{\bm \theta}]$ is used in $P(Y^{(i)} |  R_i,  {\bm \tau}_i, {\bm q}_i, {\bm \mu}, \mathcal{V}[{\bm \theta}])$ implies that the algorithm would use the black-box $\mathcal{V}$ to evaluate the the hyper-molecule at some points in order to produce an image using the algorithm's own imaging models. In fact, this can be numerically inaccurate and computationally expensive without certain assumptions on the structure of $\mathcal{V}$. 
It is therefore useful to implement efficient functions that produce projections of the hyper-molecule that are consistent with the model implemented internally in the black box $\mathcal{V}$.
In addition,  algorithms such as MALA and HMC benefit from models that can be differentiated, such that the gradients of the log-likelihood with respect to ${\bm \theta}$ and other variables such as $ R_i$ and  ${\bm \tau}_i$ are available to the algorithm.
In our implementation, such a module computes $\mathrm{log}(P(Y^{(i)} |  R_i,  {\bm \tau}_i, {\bm q}_i, {\bm \mu}, \mathcal{V}[{\bm \theta}]))$ (i.e., the comparison to the particle image is done internally in the module). Our current implementation computes gradients only with respect to  ${\bm \theta}$.

These considerations highlight the fact that complete decoupling of the hyper-molecule model from other components may present a trade-off between generality and efficient implementation considerations.

\section{Implementation and Numerical Results}\label{sec:res}

 In this section we discuss a prototype constructed for the recovery of hyper-molecules based on the ideas presented in this paper, and present the results of experiments with synthetic data. 
This implementation extends an early simplified prototype and a simpler model that did not take shifts and in-plane filters into account and allowed only 1-D non-localized heterogeneity; that prototype was not based on  MCMC.
The earlier prototype  is discussed in more detail in an earlier technical report \cite{lederman2017continuously}. Examples of objects reconstructed with the earlier prototype are presented in Figure \ref{fig:toy:cat} (bottom).

The current prototype implements simple composite hyper-molecules (see Section \ref{sec:compositehyperobj}); 
the user can define the number and positions of heterogeneous components of the hyper-molecule.
Each component can be defined to be rigid, or heterogeneous with a 1-D or 2-D state space.
Finally, the user can define components that share the same parameters, but not the same state; 
in the pretzel example, the two arms are modeled using the same coefficients $\theta$,
but in each image each arm can be in a different state.
Each object is represented using 3-D generalized prolate spheroidal functions, 
which are the optimal basis for representing objects that are as concentrated as possible in the spatial domain and in the frequency domain (as close as possible to ``compactly supported and band-limited''); for more details see \cite{lederman2017prolate}. These 3-D basis functions are multiplied by 1-D or 2-D cosines and sines to produce higher-dimensional components.

The MCMC algorithm implements MALA steps for updating the coefficients ${\bm \theta}$ of the hyper-molecule, 
and simpler MH steps (random perturbation of the variables to propose new values) for updating the viewing direction, state, in-plane shift, and contrast of each particle image.
We are working on implementing MALA and HMC for additional variables. 
The algorithm has a second mode, provided as a crude approximation of MCMC, where in each iteration, only a subset of the particle image variables (viewing direction, state, etc.) are updated (using a MH step for each particle image); the hyper-molecule is updated using a gradient step, based only on the subset of particle images considered in that iteration.
The prototype was implemented in Matlab.

We generated a dataset of 20,000 synthetic pretzel images (synthetic model described in Section \ref{sec:toy:Pretzel}), $151 \times 151$ pixels each, 
at an SNR of  $1/30$, and included effects of in-plane shifts and CTF. 
First, we set up a homogeneous model in the algorithm (the dataset is heterogeneous), and ran the algorithm with random initial viewing directions and in-plane shifts, and with the initial model set to zero everywhere. 
This run produces an initial alignment of the viewing directions.
Next, we set up the model depicted in Figure \ref{fig:toy:pretzel:parts}, with a rigid object supported in the yellow sphere, 
and two heterogeneous regions, each supported in one of the other spheres. The two heterogeneous regions are identical components (share coefficients, but shifted and rotated with respect to one another),
but each of them can appear in a different state in each particle image. The state variable is initialized uniformly at random. 
The algorithm starts with a low-frequency representation of hyper-molecule (initialized again to zero), then gradually increases the frequencies allowed in the representation; 
the gradual increase in frequency of the representation of 3-D density functions is common practice in cryo-EM \cite{scheres-relion,marina}, which is generalized here to gradual increase in the 
frequencies allowed in the state variable.
The processing requires  5 days, using a server equipped with a E5-2680 CPU and one NVIDIA Tesla P100 GPU with 16 GB of RAM. The results are presented in Figure \ref{fig:toy:pretzel:exp}(bottom).

\section{Discussion and Future Work}\label{sec:discussion}

The main goal of this paper is to introduce the idea of hyper-molecules as high-dimensional representations of 3-D molecules at all their conformations; this idea is applicable to other inverse problem such as CT. 
In addition to the generalization of 3-D molecules to hyper-molecules, we generalize the Bayesian formulation of cryo-EM to a Bayesian formulation of continuous heterogeneity in cryo-EM.
Compared to existing work on representing molecules in a small number of discrete conformations, hyper-molecules provide a way of describing a continuum of structure and the relations between states. 

These higher dimensional objects can be represented as generic high-dimensional functions, but we discuss statistical and computational motivations to introduce additional models of hyper-molecules, that describe more specific objects, when prior knowledge is available.
We also discuss an MCMC framework which overcomes some of the technical computational difficulties in each iteration of current algorithms in the more general settings that we propose, and we note additional benefits of this framework in characterizing the uncertainty in solutions.
Furthermore, we note that the MCMC framework provides a natural connection to atomic structures and other experiment modalities, demonstrated for example in \cite{habeck2017bayesian}, which uses a density map produced from a cryo-EM experiment together with physical models and other modalities.

Ultimately, the goal of this line of work is to provide a highly customizable framework for encoding prior knowledge about complex molecules and to find a practical trade-off between the bias that can be introduced by assumptions and the realistic constraints on the amount of data that can be collected. We envision this framework as a combination of imaging modules for modeling hyper-molecules adapted to fast computation of projection images and to computing gradients with respect to variables such as the viewing direction and model coefficients. 
Such modules will be used in a framework inspired by TensorFlow \cite{tensorflow2015-whitepaper}, PyTorch \cite{paszke2017automatic} (both designed primarily for deep learning) and Edward \cite{tran2016edward,edward2017}, which allow to construct modules analogous to the black-box modules discussed in this paper, with more focus on imaging as in ODL \cite{adler2017odl}.
Ideally, a wide array of general purpose tools and algorithms constructed for optimization, Bayesian inference, deep learning and imaging could be used together with this framework. 
However, the large scale of the cryo-EM problem and various properties of the problem require a more specialized framework and flexibility in solver strategies; for example, the memory management in software designed for deep learning is often optimized for small batches, whereas in some implementations of imaging algorithms there are computational advantages in working with very large batches. Another example is the update of in-plane shift variables, which can be performed without recomputing the entire image.
Among other things, a speedup may be obtained by simultaneously computing cross-correlations for multiple in-plane rotations using the recently proposed method of \cite{RanganEtAl2019}.
We demonstrated the ideas in this paper in a prototype implementation; we are currently building the next prototype, which will be more customizable and scalable.

Our reference to tools such as TensorFlow, PyTorch and Edward demonstrates that the lines between optimization, stochastic optimization, MCMC and other algorithms are not entirely rigid, in the sense that modules used in one framework can be used in some other frameworks. 
We expect to experiment with other algorithms for initialization of MCMC and approximation of steps, and to examine additional Bayesian inference algorithms. Indeed, we have already experimented with  expectation-maximization algorithms to initialize crude viewing directions in cryo-EM data and with SGD hybrids for approximating MCMC steps.

In the following sections we briefly comment on some additional aspects of the problem.

\subsection{The Homogeneous Case, Discrete Heterogeneity, and Continuous Heterogeneity}

In many cases, molecules appear mainly in a discrete set of conformations that are very similar to one another. 
While we mainly discuss continuous heterogeneity in the paper, the framework proposed here applies to the discrete case (or mixtures of discrete and continuous heterogeneity in different regions) with few changes (for example, the basis functions used to capture the variability as a function of the heterogeneity parameter ${\bm{\tau}}$ can be replaced by the Haar basis). 
Hyper-molecules, composite hyper-molecules and the algorithms discussed here are advantageous in the discrete case as well: they allow to use the similarity between different conformations, and they allow to decompose the heterogeneity to local heterogeneity in different regions.

More generally, we hope that a generic Bayesian framework could also be used to study more elaborate models for imaging and experiment latent variables even in the homogeneous case.

\subsection{Ambiguity}\label{sec:amb}

We note that even in the classic cryo-EM problem, certain ambiguities emerge in the macro-molecules that are recovered: any result has ``equivalent'' results that are identical up to global rotation, shifts and reflection. 
Naturally, hyper-molecules have similar ambiguities. Since hyper-objects generalize the spatial coordinates and in many ways treat the state parameters in the same way as they treat the spatial coordinate, 
one may expect a generalized form of ambiguity to appear. Indeed, there is ambiguity in how the molecules in different states are aligned with respect to each other and ambiguity in the parameterization of the state space. 
These ambiguities are reduced by regularization or priors, or when when the model contains rigid components that align other components.

One such effect can be observed in the cat example in Figure \ref{fig:toy:cat}, where the recovered cats are aligned slightly differently with respect to each other compared to the original cats (the change in alignment is continuous, so the ``movie of cat'' is still continuous). Of course, our original alignment was arbitrary, so the algorithm's choice is no better or worse than ours, but it is better suited to the limited degree polynomials we allowed the algorithm to use to represent these recovered cats.

\section{Conclusions}\label{sec:conclusions}

A mathematical formulation and a Bayesian formulation has been presented for the modeling of continuously heterogeneous molecular structures.
This formulation ``hyper-molecules'' and its generalizations allow to model generic heterogeneous molecules or to encode structural constraints and priors where these are available or required for practical reasons. 

In addition, we presented a computational framework based on MCMC for the recovery of hyper-molecules from cryo-EM data. 
This framework addresses some of the computational challenges associated with generalizing existing popular algorithms to the cryo-EM problem.
In particular, it bypasses the computationally intensive estimation of the conditional distribution of variables such as the viewing direction 
of each particle image at each iteration of expectation-maximization, which would become more computationally demanding if additional state variables are introduced in the case of continuous heterogeneity. This framework also offers a natural way to incorporate elaborate black-box models that researchers can customize for their needs and a tool for studying the uncertainty in solutions. 

The ideas presented in this paper have been demonstrated in a prototype implementation applied to synthetic data. 
Work on real datasets will be discussed separately. 
More scalable implementations are being constructed for more generic models, larger datasets, and more efficient computation.

\section*{Acknowledgments}

The authors would like to thank Fred Sigworth and Tejal Bhamre for their help.

A. Singer was partially supported by NIGMS Award Number R01GM090200, AFOSR FA955017-1-0291, Simons Investigator Award, the Moore Foundation Data-Driven Discovery Investigator Award, and NSF BIGDATA Award IIS-1837992. These awards also partially supported R. R. Lederman at Princeton University.
The Flatiron Institute is a division of the Simons Foundation.

\section*{References}

\bibliography{bib10}{}

\begin{thebibliography}{10}

\bibitem{kuhlbrandt}
W.~K\"{u}hlbrandt.
\newblock The resolution revolution.
\newblock {\em Science}, 343(6178):1443--1444, 2014.

\bibitem{smith2014beyond}
Martin~TJ Smith and John~L Rubinstein.
\newblock Beyond blob-ology.
\newblock {\em Science}, 345(6197):617--619, 2014.

\bibitem{cheng-trpv1}
Maofu Liao, Erhu Cao, David Julius, and Yifan Cheng.
\newblock Structure of the {TRPV1} ion channel determined by electron
  cryo-microscopy.
\newblock {\em Nature}, 504(7478):107--112, 2013.

\bibitem{amunts2014structure}
Alexey Amunts, Alan Brown, Xiao-chen Bai, Jose~L. Ll{\'a}cer, Tanweer Hussain,
  Paul Emsley, Fei Long, Garib Murshudov, Sjors H.~W. Scheres, and
  V.~Ramakrishnan.
\newblock Structure of the yeast mitochondrial large ribosomal subunit.
\newblock {\em Science}, 343(6178):1485--1489, 2014.

\bibitem{bartesaghi20152}
Alberto Bartesaghi, Alan Merk, Soojay Banerjee, Doreen Matthies, Xiongwu Wu,
  Jacqueline~LS Milne, and Sriram Subramaniam.
\newblock 2.2 {{\AA}} resolution cryo-{EM} structure of $\beta$-galactosidase
  in complex with a cell-permeant inhibitor.
\newblock {\em Science}, 348(6239):1147--1151, 2015.

\bibitem{scheres-relion}
S.~Scheres.
\newblock {RELION}: {I}mplementation of a {B}ayesian approach to cryo-{EM}
  structure determination.
\newblock {\em J. Struct. Biol.}, 180(3):519--530, 2012.

\bibitem{brubaker}
Ali Punjani, John~L Rubinstein, David~J Fleet, and Marcus~A Brubaker.
\newblock {cryoSPARC}: algorithms for rapid unsupervised cryo-{EM} structure
  determination.
\newblock {\em Nat. Methods}, 14(3):290--296, 2017.

\bibitem{eman2}
Guang Tang, Liwei Peng, Philip~R Baldwin, Deepinder~S Mann, Wen Jiang, Ian
  Rees, and Steven~J Ludtke.
\newblock {EMAN2}: an extensible image processing suite for electron
  microscopy.
\newblock {\em J. Struct. Biol.}, 157(1):38--46, 2007.

\bibitem{van2006four}
Marin van Heel, Rodrigo Portugal, A~Rohou, C~Linnemayr, C~Bebeacua, R~Schmidt,
  T~Grant, and M~Schatz.
\newblock Four-dimensional cryo-electron microscopy at quasi-atomic resolution:
  Imagic 4d.
\newblock {\em International Tables for Crystallography}, pages 624--628, 2006.

\bibitem{de2013xmipp}
JM~De~la Rosa-Trev{\'\i}n, J~Ot{\'o}n, R~Marabini, A~Zaldivar, J~Vargas,
  JM~Carazo, and COS Sorzano.
\newblock Xmipp 3.0: an improved software suite for image processing in
  electron microscopy.
\newblock {\em J. Struct. Biol.}, 184(2):321--328, 2013.

\bibitem{frealign}
Nikolaus Grigorieff.
\newblock {FREALIGN}: {H}igh-resolution refinement of single particle
  structures.
\newblock {\em J. Struct. Biol.}, 157(1):117 -- 125, 2007.

\bibitem{liu2017structural}
Daifei Liu, Xueqi Liu, Zhiguo Shang, and Charles~V Sindelar.
\newblock Structural basis of cooperativity in kinesin revealed by {3D}
  reconstruction of a two-head-bound state on microtubules.
\newblock {\em eLife}, 6:e24490, 2017.

\bibitem{dolino2016conformational}
Drew~M Dolino, Soheila~Rezaei Adariani, Sana~A Shaikh, Vasanthi Jayaraman, and
  Hugo Sanabria.
\newblock Conformational selection and submillisecond dynamics of the
  ligand-binding domain of the n-methyl-d-aspartate receptor.
\newblock {\em Journal of Biological Chemistry}, 291(31):16175--16185, 2016.

\bibitem{nogales2016development}
Eva Nogales.
\newblock The development of cryo-{EM} into a mainstream structural biology
  technique.
\newblock {\em Nat. Methods}, 13(1):24--27, 2016.

\bibitem{glaeser2016good}
Robert~M Glaeser.
\newblock How good can cryo-{EM} become?
\newblock {\em Nat. Methods}, 13(1):28--32, 2016.

\bibitem{scheres}
Sjors~HW Scheres.
\newblock A {B}ayesian view on cryo-{EM} structure determination.
\newblock {\em J. Mol. Biol.}, 415(2):406--418, 2012.

\bibitem{sorzano2019survey}
COS Sorzano, A~Jim{\'e}nez, J~Mota, JL~Vilas, D~Maluenda, M~Mart{\'\i}nez,
  E~Ram{\'\i}rez-Aportela, T~Majtner, J~Segura, R~S{\'a}nchez-Garc{\'\i}a,
  et~al.
\newblock Survey of the analysis of continuous conformational variability of
  biological macromolecules by electron microscopy.
\newblock {\em Acta Crystallographica Section F: Structural Biology
  Communications}, 75(1):19--32, 2019.

\bibitem{low2003method}
Daniel~A Low, Michelle Nystrom, Eugene Kalinin, Parag Parikh, James~F Dempsey,
  Jeffrey~D Bradley, Sasa Mutic, Sasha~H Wahab, Tareque Islam, Gary
  Christensen, et~al.
\newblock A method for the reconstruction of four-dimensional synchronized {CT}
  scans acquired during free breathing.
\newblock {\em Medical Physics}, 30(6):1254--1263, 2003.

\bibitem{kimanius}
Dari Kimanius, Bj{\"o}rn~O Forsberg, Sjors~HW Scheres, and Erik Lindahl.
\newblock Accelerated cryo-{EM} structure determination with parallelisation
  using {GPUs} in {RELION}-2.
\newblock {\em eLife}, 5, nov 2016.

\bibitem{zivanov2018new}
Jasenko Zivanov, Takanori Nakane, Bj{\"o}rn~O Forsberg, Dari Kimanius, Wim~JH
  Hagen, Erik Lindahl, and Sjors~HW Scheres.
\newblock New tools for automated high-resolution cryo-em structure
  determination in relion-3.
\newblock {\em eLife}, 7:e42166, 2018.

\bibitem{lederman2017continuously}
Roy~R. Lederman and Amit Singer.
\newblock Continuously heterogeneous hyper-objects in cryo-{EM} and {3-D}
  movies of many temporal dimensions.
\newblock {\em arXiv preprint arXiv:1704.02899}, 2017.

\bibitem{lederman2017prolate}
Roy~R. Lederman.
\newblock Numerical algorithms for the computation of generalized prolate
  spheroidal functions.
\newblock {\em arXiv preprint arXiv:1710.02874}, 2017.

\bibitem{kawabata2008multiple}
Takeshi Kawabata.
\newblock Multiple subunit fitting into a low-resolution density map of a
  macromolecular complex using a {G}aussian mixture model.
\newblock {\em Biophysical Journal}, 95(10):4643--4658, 2008.

\bibitem{lederman2019representation}
Roy~R Lederman and Amit Singer.
\newblock A representation theory perspective on simultaneous alignment and
  classification.
\newblock {\em Applied and Computational Harmonic Analysis}, 2019.

\bibitem{frank}
J.~Frank.
\newblock {\em Three-dimensional electron microscopy of macromolecular
  assemblies}.
\newblock Academic Press, 2006.

\bibitem{sigworth-review}
Fred~J. Sigworth.
\newblock {Principles of cryo-EM single-particle image processing}.
\newblock {\em Microscopy}, 65(1):57--67, 12 2015.

\bibitem{cheng2015primer}
Yifan Cheng, Nikolaus Grigorieff, Pawel~A. Penczek, and Thomas Walz.
\newblock A primer to single-particle cryo-electron microscopy.
\newblock {\em Cell}, 161(3):438--449, 2015.

\bibitem{milne2013cryo}
Jacqueline~LS Milne, Mario~J Borgnia, Alberto Bartesaghi, Erin~EH Tran,
  Lesley~A Earl, David~M Schauder, Jeffrey Lengyel, Jason Pierson, Ardan
  Patwardhan, and Sriram Subramaniam.
\newblock Cryo-electron microscopy--{A} primer for the non-microscopist.
\newblock {\em FEBS Journal}, 280(1):28--45, 2013.

\bibitem{vinothkumar2016single}
Kutti~R Vinothkumar and Richard Henderson.
\newblock Single particle electron cryomicroscopy: {T}rends, issues and future
  perspective.
\newblock {\em Q. Rev. Biophys.}, 49, 2016.

\bibitem{fred-ml}
Fred~J. Sigworth.
\newblock A maximum-likelihood approach to single-particle image refinement.
\newblock {\em J. Struct. Biol.}, 122(3):328--339, 1998.

\bibitem{sigworth2010chapter}
Fred~J Sigworth, Peter~C Doerschuk, Jose-Maria Carazo, and Sjors~HW Scheres.
\newblock Chapter ten---an introduction to maximum-likelihood methods in
  cryo-{EM}.
\newblock {\em Methods Enzymol.}, 482:263--294, 2010.

\bibitem{punjani2016building}
Ali Punjani, Marcus Brubaker, and David Fleet.
\newblock Building proteins in a day: Efficient {3D} molecular structure
  estimation with electron cryomicroscopy.
\newblock {\em IEEE Trans. Pattern Anal. Mach. Intell.}, 2016.

\bibitem{shatsky}
M.~Shatsky, R.~Hall, E.~Nogales, J.~Malik, and S.~Brenner.
\newblock Automated multi-model reconstruction from single-particle electron
  microscopy data.
\newblock {\em J. Struct. Biol.}, 170(1):98--108, 2010.

\bibitem{singer2010detecting}
Amit Singer, Ronald~R Coifman, Fred~J Sigworth, David~W Chester, and Yoel
  Shkolnisky.
\newblock Detecting consistent common lines in cryo-{EM} by voting.
\newblock {\em J. Struct. Biol.}, 169(3):312--322, 2010.

\bibitem{shkolnisky2012viewing}
Yoel Shkolnisky and Amit Singer.
\newblock Viewing direction estimation in cryo-{EM} using synchronization.
\newblock {\em SIAM J. Imaging Sci.}, 5(3):1088--1110, 2012.

\bibitem{bandeira2015non}
Afonso~S Bandeira, Yutong Chen, and Amit Singer.
\newblock Non-unique games over compact groups and orientation estimation in
  cryo-{EM}.
\newblock {\em arXiv preprint arXiv:1505.03840}, 2015.

\bibitem{joubert2015bayesian}
Paul Joubert and Michael Habeck.
\newblock Bayesian inference of initial models in cryo-electron microscopy
  using pseudo-atoms.
\newblock {\em Biophysical Journal}, 108(5):1165--1175, 2015.

\bibitem{dashti}
Ali Dashti, Peter Schwander, Robert Langlois, Russell Fung, Wen Li, Ahmad
  Hosseinizadeh, Hstau~Y. Liao, Jesper Pallesen, Gyanesh Sharma, Vera~A.
  Stupina, Anne~E. Simon, Jonathan~D. Dinman, Joachim Frank, and Abbas Ourmazd.
\newblock Trajectories of the ribosome as a {B}rownian nanomachine.
\newblock {\em Proc. Natl. Acad. Sci. U.S.A.}, 111(49):17492--17497, 2014.

\bibitem{schwander2014conformations}
P~Schwander, R~Fung, and A~Ourmazd.
\newblock Conformations of macromolecules and their complexes from
  heterogeneous datasets.
\newblock {\em Phil. Trans. R. Soc. B}, 369(1647):20130567, 2014.

\bibitem{frank2016continuous}
Joachim Frank and Abbas Ourmazd.
\newblock Continuous changes in structure mapped by manifold embedding of
  single-particle data in cryo-{EM}.
\newblock {\em Methods}, 100:61--67, 2016.

\bibitem{nakane2018characterisation}
Takanori Nakane, Dari Kimanius, Erik Lindahl, and Sjors~HW Scheres.
\newblock Characterisation of molecular motions in cryo-{EM} single-particle
  data by multi-body refinement in {RELION}.
\newblock {\em eLife}, 7:e36861, 2018.

\bibitem{wong2014cryo}
Wilson Wong, Xiao-chen Bai, Alan Brown, Israel~S Fernandez, Eric Hanssen,
  Melanie Condron, Yan~Hong Tan, Jake Baum, and Sjors~HW Scheres.
\newblock Cryo-{EM} structure of the {P}lasmodium falciparum 80s ribosome bound
  to the anti-protozoan drug emetine.
\newblock {\em eLife}, 3:e03080, 2014.

\bibitem{zhou2015cryo}
Qiang Zhou, Xuan Huang, Shan Sun, Xueming Li, Hong-Wei Wang, and Sen-Fang Sui.
\newblock Cryo-em structure of snap-snare assembly in 20s particle.
\newblock {\em Cell Research}, 25(5):551, 2015.

\bibitem{bai2015sampling}
Xiao-chen Bai, Eeson Rajendra, Guanghui Yang, Yigong Shi, and Sjors~HW Scheres.
\newblock Sampling the conformational space of the catalytic subunit of human
  $\gamma$-secretase.
\newblock {\em eLife}, 4:e11182, 2015.

\bibitem{ilca2015localized}
Serban~L Ilca, Abhay Kotecha, Xiaoyu Sun, Minna~M Poranen, David~I Stuart, and
  Juha~T Huiskonen.
\newblock Localized reconstruction of subunits from electron cryomicroscopy
  images of macromolecular complexes.
\newblock {\em Nat. Commun.}, 6:8843, 2015.

\bibitem{anden2018structural}
Joakim And{\'{e}}n and Amit Singer.
\newblock {Structural Variability from Noisy Tomographic Projections}.
\newblock {\em SIAM J. Imaging Sci.}, 11(2):1441--1492, jan 2018.

\bibitem{tama2002exploring}
Florence Tama, Willy Wriggers, and Charles~L Brooks~III.
\newblock Exploring global distortions of biological macromolecules and
  assemblies from low-resolution structural information and elastic network
  theory.
\newblock {\em J. Mol. Biol.}, 321(2):297--305, 2002.

\bibitem{jin2014iterative}
Qiyu Jin, Carlos Oscar~S. Sorzano, Jos\'{e}~Miguel {de~la}~{Rosa-Trev\'{i}n},
  Jos\'{e}~Rom\'{a}n {Bilbao-Castro}, Rafael {N\'{u}\~{n}ez-Ram\'{i}rez}, Oscar
  Llorca, Florence Tama, and Slavica Joni\'{c}.
\newblock Iterative elastic {3D}-to-{2D} alignment method using normal modes
  for studying structural dynamics of large macromolecular complexes.
\newblock {\em Structure}, 22(3):496--506, 2014.

\bibitem{brooks2011handbook}
Steve Brooks, Andrew Gelman, Galin Jones, and Xiao-Li Meng.
\newblock {\em Handbook of {M}arkov chain {M}onte {C}arlo}.
\newblock CRC press, 2011.

\bibitem{welling2011bayesian}
Max Welling and Yee~W. Teh.
\newblock Bayesian learning via stochastic gradient {L}angevin dynamics.
\newblock In {\em Proc. ICML}, pages 681--688, 2011.

\bibitem{LiuFrank1995}
Weiping Liu and Joachim Frank.
\newblock Estimation of variance distribution in three-dimensional
  reconstruction. {I}. {T}heory.
\newblock {\em J. Opt. Soc. Am. A}, 12(12):2615--2627, Dec 1995.

\bibitem{Penczek2002}
P.~A. Penczek.
\newblock Variance in three-dimensional reconstructions from projections.
\newblock In {\em Proc. ISBI}, pages 749--752, 2002.

\bibitem{PenczekEtal2006}
Pawel~A. Penczek, Chao Yang, Joachim Frank, and Christian~M.T. Spahn.
\newblock Estimation of variance in single-particle reconstruction using the
  bootstrap technique.
\newblock {\em J. Struct. Biol.}, 154(2):168--183, 2006.

\bibitem{LiaoFrank2010}
H.~Liao and J.~Frank.
\newblock Classification by bootstrapping in single particle methods.
\newblock In {\em Proc. ISBI}, pages 169--172. IEEE, April 2010.

\bibitem{PenczekKimmelSpahn2011}
P.~Penczek, M.~Kimmel, and C.~Spahn.
\newblock Identifying conformational states of macromolecules by eigen-analysis
  of resampled cryo-{EM} images.
\newblock {\em Structure}, 19(11):1582--1590, 2011.

\bibitem{AndenKatsevichSinger2015}
J.~And\'{e}n, E.~Katsevich, and A.~Singer.
\newblock Covariance estimation using conjugate gradient for {3D}
  classification in cryo-em.
\newblock In {\em Proc. ISBI}, pages 200--204, April 2015.

\bibitem{AndenSinger2018}
Joakim And{\'{e}}n and Amit Singer.
\newblock Structural variability from noisy tomographic projections.
\newblock {\em SIAM J. Imaging Sci.}, 11(2):1441--1492, may 2018.

\bibitem{marina}
Alex Barnett, Leslie Greengard, Andras Pataki, and Marina Spivak.
\newblock Rapid solution of the cryo-{EM} reconstruction problem by frequency
  marching.
\newblock {\em SIAM J. Imaging Sci.}, 10(3):1170--1195, 2017.

\bibitem{habeck2017bayesian}
Michael Habeck.
\newblock Bayesian modeling of biomolecular assemblies with cryo-{EM} maps.
\newblock {\em Frontiers in Molecular Biosciences}, 4:15, 2017.

\bibitem{tensorflow2015-whitepaper}
Mart\'{\i}n Abadi, Ashish Agarwal, Paul Barham, Eugene Brevdo, Zhifeng Chen,
  Craig Citro, Greg~S. Corrado, Andy Davis, Jeffrey Dean, Matthieu Devin,
  Sanjay Ghemawat, Ian Goodfellow, Andrew Harp, Geoffrey Irving, Michael Isard,
  Yangqing Jia, Rafal Jozefowicz, Lukasz Kaiser, Manjunath Kudlur, Josh
  Levenberg, Dandelion Man\'{e}, Rajat Monga, Sherry Moore, Derek Murray, Chris
  Olah, Mike Schuster, Jonathon Shlens, Benoit Steiner, Ilya Sutskever, Kunal
  Talwar, Paul Tucker, Vincent Vanhoucke, Vijay Vasudevan, Fernanda Vi\'{e}gas,
  Oriol Vinyals, Pete Warden, Martin Wattenberg, Martin Wicke, Yuan Yu, and
  Xiaoqiang Zheng.
\newblock {TensorFlow}: Large-scale machine learning on heterogeneous systems,
  2015.
\newblock Software available from tensorflow.org.

\bibitem{paszke2017automatic}
Adam Paszke, Sam Gross, Soumith Chintala, Gregory Chanan, Edward Yang, Zachary
  DeVito, Zeming Lin, Alban Desmaison, Luca Antiga, and Adam Lerer.
\newblock Automatic differentiation in {PyTorch}.
\newblock 2017.

\bibitem{tran2016edward}
Dustin Tran, Alp Kucukelbir, Adji~B. Dieng, Maja Rudolph, Dawen Liang, and
  David~M. Blei.
\newblock {Edward: A library for probabilistic modeling, inference, and
  criticism}.
\newblock {\em arXiv preprint arXiv:1610.09787}, 2016.

\bibitem{edward2017}
Dustin Tran, Matthew~D. Hoffman, Rif~A. Saurous, Eugene Brevdo, Kevin Murphy,
  and David~M. Blei.
\newblock Deep probabilistic programming.
\newblock In {\em Proc. ICLR}, 2017.

\bibitem{adler2017odl}
Jonas Adler, Holger Kohr, and Ozan {\"O}ktem.
\newblock {ODL}---a {P}ython framework for rapid prototyping in inverse
  problems.
\newblock Technical report, Royal Institute of Technology, 2017.

\bibitem{RanganEtAl2019}
Aaditya Rangan, Marina Spivak, Joakim And\'en, and Alex Barnett.
\newblock Fast rigid image alignment by {F}ourier--{B}essel factorization of
  inner products.
\newblock Submitted to Inverse Problems, 2019.

\end{thebibliography}
\bibliographystyle{unsrt}

\end{document}